\documentclass[10pt,twocolumn,letterpaper]{article}

\usepackage{iccv}              
\usepackage[accsupp]{axessibility}  

\usepackage{float}
\usepackage{amsmath}

\usepackage[utf8]{inputenc} 
\usepackage[T1]{fontenc}    
\usepackage{url}            
\usepackage{booktabs}       
\usepackage{amsfonts}       
\usepackage{nicefrac}       
\usepackage{microtype}      
\usepackage{xcolor}         

\usepackage{algorithm,algpseudocode}
\usepackage{algorithmicx}

\usepackage{times}
\usepackage{epsfig}
\usepackage{graphicx}
\usepackage{amssymb}
\usepackage{color}
\usepackage{multirow}
\usepackage{tabularx}
\usepackage{caption}
\usepackage{silence}
\usepackage[accsupp]{axessibility}

\newcommand{\modelname}{\textbf{HypDAE}}

\definecolor{iccvblue}{rgb}{0.21,0.49,0.74}
\usepackage[pagebackref,breaklinks,colorlinks,allcolors=iccvblue]{hyperref}

\def\paperID{11469} 
\def\confName{ICCV}
\def\confYear{2025}

\title{\textsc{HypDAE}: Hyperbolic Diffusion Autoencoders for \\Hierarchical Few-shot Image Generation}

\author{
Lingxiao Li\textsuperscript{\textnormal{1, 2}} \quad Kaixuan Fan\textsuperscript{\textnormal{1}} \quad Boqing Gong\textsuperscript{\textnormal{2}}\thanks{Corresponding authors} \quad Xiangyu Yue\textsuperscript{\textnormal{$1 *$}}\\ \\\textsuperscript{1} MMLab, The Chinese University of Hong Kong \quad
\textsuperscript{2} Boston University\\
{\tt\small \{lxli, bgong\}@bu.edu, kxfan127@gmail.com, xyyue@ie.cuhk.edu.hk}
}

\begin{document}
\maketitle
\begin{abstract}
   Few-shot image generation aims to generate diverse and high-quality images for an unseen class given only a few examples in that class. A key challenge in this task is balancing category consistency and image diversity, which often compete with each other. Moreover, existing methods offer limited control over the attributes of newly generated images. In this work, we propose Hyperbolic Diffusion Autoencoders (HypDAE), a novel approach that operates in hyperbolic space to capture hierarchical relationships among images from seen categories. By leveraging pre-trained foundation models, HypDAE generates diverse new images for unseen categories with exceptional quality by varying stochastic subcodes or semantic codes. Most importantly, the hyperbolic representation introduces an additional degree of control over semantic diversity through the adjustment of radii within the hyperbolic disk. Extensive experiments and visualizations demonstrate that HypDAE significantly outperforms prior methods by achieving a better balance between preserving category-relevant features and promoting image diversity with limited data. Furthermore, HypDAE offers a highly controllable and interpretable generation process. Code is available at: \href{https://github.com/lingxiao-li/HypDAE}{https://github.com/lingxiao-li/HypDAE}.
\end{abstract}    
\section{Introduction}
\label{sec:intro}

Generative models~\cite{Song21Scorebased, Ho20DDPM, Rombach22ldm, Ramesh22DALLE2} have succeeded in generating high-fidelity and realistic images, partially thanks to a large volume of high-quality data for model training.
However, with the widespread presence of long-tail distributions and data imbalances across image categories~\cite{Hong20F2}, there are many scenarios in the real world where it is impossible to collect sufficient samples of certain categories for model training. 
It is difficult for generative models trained on well-sampled categories to generate realistic and diverse images for a novel category given only a few examples. This challenging task is known as few-shot image generation, which aims to synthesize images that preserve the category-level identity of the limited input samples~\cite{Clouâtre19, Hong20, Hong20F2, Hong20Match, Hong22, Hong22Delta, Ding22, Li23HAE, Ding23}.


\begin{figure}[t]
  \centering
  \setlength{\abovecaptionskip}{0.2cm}   
  \setlength{\belowcaptionskip}{-0.3cm}   
   \includegraphics[width=0.95\linewidth]{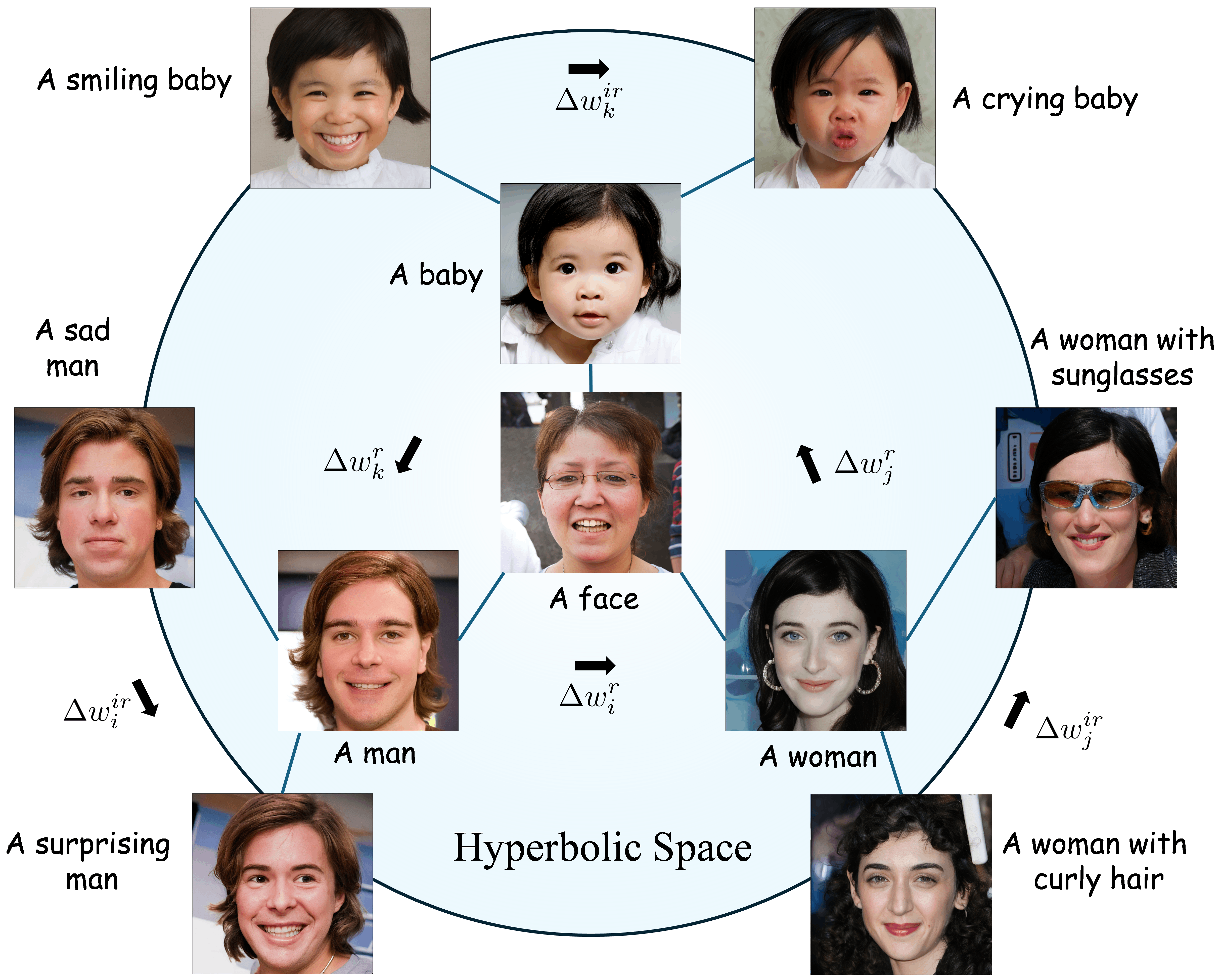}
    \vspace{-0.1cm}
   \caption{\textbf{Illustration of hierarchical text-image representation in hyperbolic space.} Hyperbolic space provides a natural and compact encoding for semantic hierarchies in large datasets. Adjusting high-level, identity-relevant attributes $\Delta \mathbf{w}^{r}$ alters an image's identity, while modifying low-level, identity-irrelevant attributes $\Delta \mathbf{w}^{ir}$ produces variations within the same identity.}
   \label{fig: Figure_1}
   \vspace{-2.0ex}
\end{figure}


Existing few-shot image generation methods are primarily GAN-based and fall into three categories: \textit{transfer-based} approaches~\cite{Clouâtre19, Liang20}, which use meta-learning or domain adaptation for cross-category generalization but often face limited transferability; \textit{fusion-based} approaches~\cite{Bartunov18, Gu21, Hong20F2, Hong20Match, yang22wavegan, Zhou24F2DGAN}, which fuse features from multiple exemplars but tend to produce outputs overly similar to the inputs; and \textit{transformation-based} approaches~\cite{Antoniou17, Hong22, Hong22Delta, Ding22, Ding23}, which apply intra-category perturbations without fine-tuning but often lack diversity. Recent diffusion-based methods~\cite{Kumari22customdiffusion, Ruiz23DreamBooth, Cao23Masactrl} for \textit{object-level personalization} focus on instance identity, require test-time fine-tuning, and depend on prompt engineering, making them incompatible with our category-level setting without textual inputs or model updates. However, all these methods operate in Euclidean feature space, which limits their ability to capture hierarchical semantic structure—a key requirement for generalizing from few examples in category-level few-shot image generation.

In contrast to prior methods, recent work such as HAE~\cite{Li23HAE} highlights the importance of modeling hierarchical structures in few-shot image generation. Similar to language~\cite{Nickel17, Tifrea19, Dhingra18}, images also exhibit semantic hierarchies~\cite{Khrulkov21, Cui23}, where each image can be viewed as a composition of attributes at different levels. As illustrated in \cref{fig: Figure_1}, high-level, identity-relevant attributes (e.g., gender or age) define the semantic core of a category, while low-level, identity-irrelevant attributes (e.g., expression or hairstyle) introduce intra-class variation. Capturing this hierarchical organization is essential for generating diverse yet category-consistent images. Hyperbolic space provides a natural embedding for such structures, as it can represent tree-like relationships with low distortion~\cite{Nickel17}. This enables infinite layers of semantics to be encoded compactly, allowing identity-preserving edits along radial directions and intra-category diversity through tangential shifts in the latent space, which facilitates a structured and interpretable latent space that supports controllable few-shot image generation.

Despite the advantages of hierarchical representation for few-shot image generations~\cite{Li23HAE}, existing methods are dominated by GAN-based approaches and face
three primary challenges: \textbf{1)} Suboptimal image quality due to the generative constraints of GANs, particularly with insufficient training data; \textbf{2)} Reduced diversity from the 1-to-1 mapping between hyperbolic and image spaces, which leads to the loss of high-frequency details when latent codes are insufficiently trained; and \textbf{3)} The necessity of labeled data for learning hierarchical latent representations. In parallel, recent advances in diffusion models~\cite{preechakul21diffae, Kumari22customdiffusion, Ruiz23DreamBooth, Ye23ip-adapter, Cao23Masactrl} enable high-quality image generation with rich, diverse details. Moreover, pre-trained foundation models (e.g., Stable Diffusion~\cite{Rombach22ldm}, CLIP~\cite{Radford21CLIP}) further support adaptation with limited data and strong generalization.

To address the challenges, we propose \textbf{Hyp}erbolic \textbf{D}iffusion \textbf{A}uto\textbf{E}ncoders (\modelname{}), a novel method for leveraging the natural suitability of hyperbolic space for hierarchical latent code manipulation and the generative ability of diffusion models for few-shot image generation. In particular, we separate the representation of given images into two subcodes~\cite{preechakul21diffae}.
Our method begins by training an image encoder to capture high-level semantic subcode, while utilizing a pre-trained Stable Diffusion (SD)~\cite{Rombach22ldm} model for decoding and modeling stochastic subcode by reversing the generative process. A hyperbolic feature encoder and decoder are then trained to map latent vectors from Euclidean space $\mathbb{R}^n$ to hyperbolic space $\mathbb{D}^n$, with classification loss ensuring hierarchical image embeddings. Finally, to reduce the data labeling cost for training a hyperbolic encoder, pseudo-labeling is enabled by zero-shot classification with pre-trained vision models.
By capturing the attribute hierarchy among images, \modelname{} generates new images through two methods: \textbf{1)} varying stochastic subcode of the Diffusion Autoencoders, and \textbf{2)} randomly shifting semantic subcode in an identity-irrelevant direction to modify category-irrelevant features. Hyperbolic space enables control over the semantic diversity of generated images by adjusting the radii in hyperbolic space. This facilitates hierarchical attribute editing for flexible, high-quality, and diverse few-shot image generation.

\vspace{0.1cm}
\noindent Our contributions are summarized as follows: 
\begin{itemize} 
    \item We introduce \modelname{}, an innovative method for few-shot image generation, which learns the hierarchical representation of images in hyperbolic space. To the best of our knowledge, \modelname{} is the first to enable diffusion models to model semantic hierarchical information in hyperbolic space for few-shot image generation. 
    \item We demonstrate that in our hyperbolic latent space, semantic hierarchical relationships among images are reflected by their distances to the center of the Poincaré disk. This allows for easy interpretation of the latent space.
    \item By leveraging pre-trained foundation models, \modelname{} generates high-quality images with rich, diverse, category-irrelevant features from a single image. Extensive experiments show that \modelname{} significantly outperforms existing few-shot image generation methods without requiring human-annotated class labels, offering state-of-the-art quality, diversity, and flexibility. This demonstrates the robustness of \modelname{} under settings with automatically generated labels, making it a strong solution for label-free few-shot image generation.
\end{itemize}
\section{Related Work}
\label{sec:related_work}

\noindent \textbf{Few-shot Image Generation.}
Given a few samples from one novel category, few-shot image generation aims to produce more new images with high quality and various diversity. Recent advances in few-shot image generation encompass various approaches. \textit{Transfer-based methods}~\cite{Clouâtre19, Liang20, zheng23lso}, utilizing meta-learning or domain adaptation on GANs, often struggle with generating realistic images. \textit{Fusion-based methods}, aligning random vectors with conditional images~\cite{Hong20Match} or framing generation as a conditional task~\cite{Gu21, Hong20F2, yang22wavegan}, generally produce outputs with limited diversity. \textit{Transformation-based approaches}~\cite{Hong22, Hong22Delta, Li23HAE}, which transform single conditional images, can lack consistency. Ding {\it et al.}~\cite{Ding22, Ding23} explored intra-category transformations as identity-irrelevant edits using a single sample.
More recently, Diffusion-based \textit{subject-driven generation}~\cite{Dong22DreamArtist, Daras22, voynov23P+, han23highly} adapts models to incorporate new concepts with limited data, exemplified by DreamBooth~\cite{Ruiz23DreamBooth}, which assigns unique identifiers to subjects using 3-5 images. This has inspired methods like Custom Diffusion~\cite{Kumari22customdiffusion} and further improvements~\cite{Han23SVDiff, Chen23SubjectDriven, Shi24InstantBooth}. However, these works focus on instance-level identity preservation, require test-time fine-tuning, and rely heavily on prompt engineering. Therefore, their objectives, problem settings, and evaluation protocols differ fundamentally from our task.

\noindent \textbf{Hyperbolic Representation Learning.}
Recent advancements have introduced hyperbolic space into deep learning~\cite{Nickel17, Nickel18, Tifrea19, Surís21, Khrulkov21}, initially in NLP for hierarchical language modeling~\cite{Nickel17, Nickel18, Tifrea19}. Riemannian optimization techniques have enabled model training within hyperbolic spaces~\cite{Bécigneul19, Bonnabel13}. Building on this, Ganea \etal~\cite{Ganea18} developed hyperbolic adaptations of core neural network tools such as multinomial logistic regression, feed-forward, and recurrent networks. Applications of hyperbolic geometry have since expanded to image~\cite{Khrulkov21, Li23HAE}, video~\cite{Surís21}, graph data~\cite{Chami19, Park21}, and 3D shape generation~\cite{Leng24HyperSDFusion}. Desai \etal~\cite{Desai23} introduced a hyperbolic contrastive model for hierarchical text-image pairing, while HAE~\cite{Li23HAE} demonstrated hyperbolic space's advantages in few-shot image generation, although GAN-based results remained limited in quality. To our knowledge, this work is the first to integrate hyperbolic representation with diffusion models for high-quality few-shot image generation.

\begin{figure}[t]
  \centering
  \setlength{\abovecaptionskip}{0.1cm}   
  \setlength{\belowcaptionskip}{-0.3cm}   
   \includegraphics[width=0.9\linewidth]{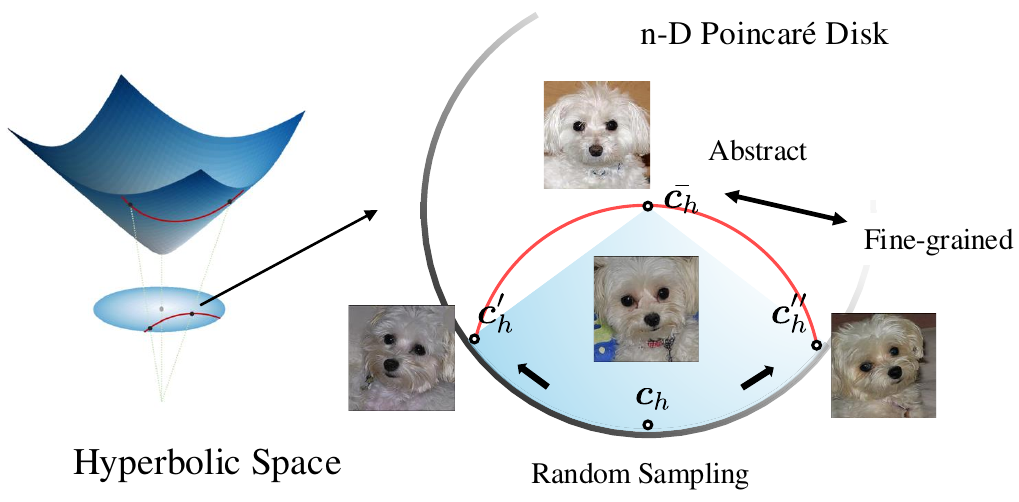}
   \caption{\textbf{Illustration of the property of hyperbolic space on the Poincaré disk.} Given two latent codes of Maltese dog $\boldsymbol{c}_{h}'$ and $\boldsymbol{c}_{h}''$ on the edge of Poincaré disk, the geodesic between these two points is the \textcolor{red}{red} curve rather than a straight line in Euclidean space. Therefore, their average latent code is calculated as $\Bar{\boldsymbol{c}_{h}}$, which is closer to the center. Thus, $\Bar{\boldsymbol{c}_{h}}$ can be viewed as the ``parent" of $\boldsymbol{c}_{h}$, $\boldsymbol{c}_{h}'$, and $\boldsymbol{c}_{h}''$. One can generate diverse images without changing the category by moving the latent code from one child to another of the same parent in the hyperbolic space.} 
   
   \label{fig: hierarchy}
   \vspace{-1.0ex}
\end{figure}

\begin{figure*}[t]
  \centering
  \setlength{\abovecaptionskip}{0.1cm}   
  \setlength{\belowcaptionskip}{-0.4cm}   
   \includegraphics[width=0.9\linewidth]{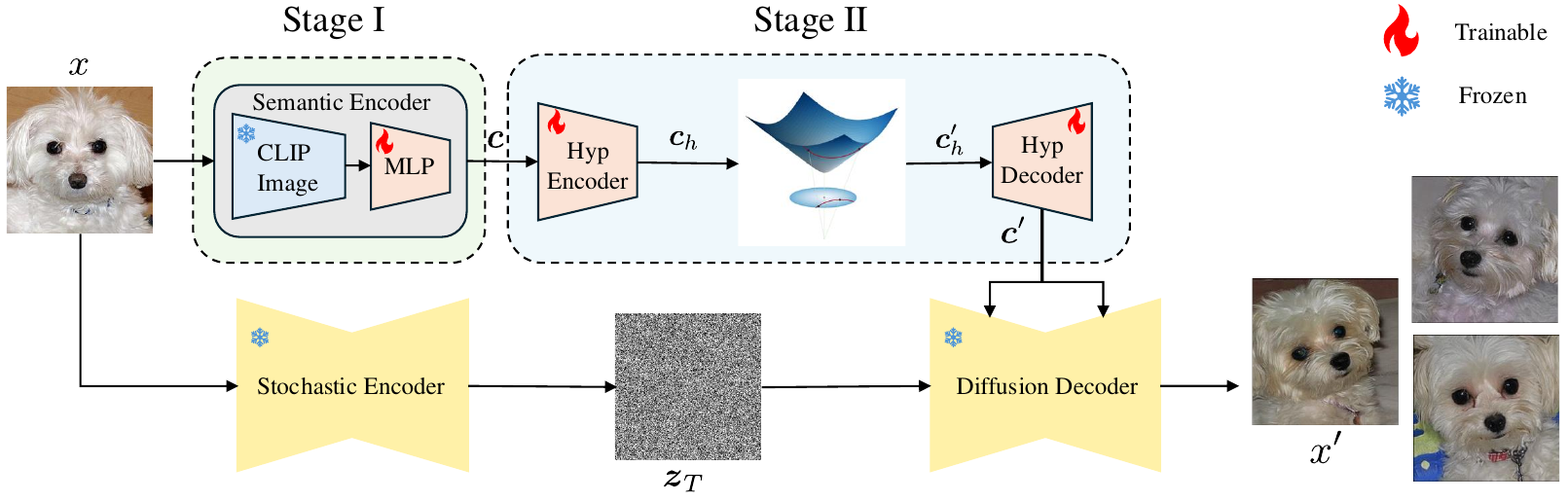}

   \caption{The overview of \modelname{}. The hyperbolic autoencoder consists of a ``semantic'' encoder that maps the reference image to the semantic code ($x \rightarrow \boldsymbol{c}$), and a stable diffusion model that acts both as a ``stochastic'' encoder ($x \rightarrow \boldsymbol{z}_T$) and a diffusion decoder ($(\boldsymbol{c}', \boldsymbol{z}_T) \rightarrow x' $). Here, $\boldsymbol{c}$ captures the high-level semantics while $\boldsymbol{z}_T$ captures low-level stochastic variations; they can be decoded back with high fidelity. The ``Hyp Encoder'' is used to project the latent code from Euclidean space $\mathbb{R}^n$ to hyperbolic space $\mathbb{D}^n$~\cite{Chami19}. The semantic code can be edited as \cref{fig: hierarchy} shows and be used as the condition for the diffusion decoder to generate diverse new images with the same category. The VAE module of SD is omitted for better illustration.}
   \label{fig: pipeline}
   \vspace{-1.0ex}
\end{figure*}
\section{Method}
\label{sec: method}
To generate diverse new images from a few reference images while preserving their identity, it is essential for our model to capture both identity-relevant features (to maintain identity) and identity-irrelevant features (to enhance diversity). To achieve this, we design a diffusion autoencoder that extracts elementary identity-relevant features via a semantic encoder and identity-irrelevant features via a stochastic encoder. Furthermore, a hyperbolic encoder-decoder is introduced to further disentangle high-order semantic features. Unlike HAE, which applies hyperbolic representations in GANs, our method integrates hyperbolic semantics into diffusion models. We introduce a stable training pipeline, and encode hyperbolic embeddings as diffusion conditions, enabling scalable and interpretable few-shot generation.

We give details of our method in the following sections. Specifically, \cref{sec: Hierarchical Learning} details the modeling of hierarchical representations, \cref{sec: netwrok} elaborates on the framework and loss functions of \modelname{}, and \cref{sec: Pseudo-Labeling} and \cref{sec: hyperbolic latent editing} discuss the pseudo-labeling process and hyperbolic latent editing algorithms, respectively.
\subsection{Preliminary}
\label{sec:preliminaries}

\noindent\textbf{Hierarchical Learning in Hyperbolic Space.}
\label{sec: Hierarchical Learning}
A key challenge in multi-level semantic editing is deriving a hierarchical representation from real images, as illustrated in \cref{fig: hierarchy}. To address this, we utilize \textit{hyperbolic} space as the latent space, given its suitability for hierarchical structures. Unlike Euclidean spaces (zero curvature) or spherical spaces (positive curvature), hyperbolic spaces exhibit negative curvature, making them ideal for modeling hierarchical data~\cite{Nickel17, Nickel18}. As a continuous analog of trees~\cite{Nickel17}, hyperbolic space enables hierarchical representation through its exponential radius growth, facilitating structured modeling across text, images, and videos~\cite{Gromov87}.

The $n$-dimensional hyperbolic space, $\mathbb{H}^n$, is a homogeneous, simply connected Riemannian manifold with constant negative sectional curvature\footnote{We set the curvature $c = -1$ in this paper.}. We use the Poincaré disk model following~\cite{Khrulkov21}, $\left(\mathbb{D}^n, g^{\mathbb{D}}\right)$, a preferred choice in gradient-based learning~\cite{Nickel17, Ganea18, Nickel18, Tifrea19, Surís21, Khrulkov21}, defined by $\mathbb{D}^n=\left\{x \in \mathbb{R}^n:\|x\|<1\right\}$ with the Riemannian metric:
\begin{equation}
    g_x^{\mathbb{D}}=\lambda_x^2 \cdot g^E,
    \label{eq:Riemannian metric}
    \end{equation}
where $\lambda_x =\frac{2}{1-\|x\|^2}$, and $g^E$ is the Euclidean metric tensor $g^E = \mathbf{I}^n$. The induced distance between two points $\mathbf{x}, \mathbf{y} \in \mathbb{D}^n$ can be defined by:
\begin{equation}
    d_{\mathbb{D}}(\mathbf{x}, \mathbf{y})=\mathrm{arccosh}\left(1+2 \frac{\|\mathbf{x}-\mathbf{y}\|^2}{\left(1-\|\mathbf{x}\|^2\right)\left(1-\|\mathbf{y}\|^2\right)}\right).
    \label{eq3-distance}
\end{equation}

A geodesic is defined as a locally minimized-length curve between two points, as illustrated in \cref{fig: hierarchy}. The midpoint of this geodesic, closer to the origin, represents the mean of two latent codes in hyperbolic space. This property ensures that the mean of two leaf embeddings is not another leaf but their hierarchical parent, as illustrated in \cref{fig: hierarchy} on a 2-D Poincaré disk~\cite{Surís21}. Embeddings near the disk's edge correspond to fine-grained images, whereas those closer to the center represent more abstract features, such as an average face.

\subsection{\textbf{Hyp}erbolic \textbf{D}iffusion \textbf{A}uto\textbf{E}ncoders (\modelname{})}
\label{sec: netwrok}
The overall pipeline of \modelname{}, illustrated in \cref{fig: pipeline}, consists of two stages:
\textbf{1)} In Stage I, a semantic encoder maps input images to meaningful latent codes, serving as conditions for the pre-trained Stable Diffusion (SD). To enhance diversity and prevent direct replication, a content bottleneck and strong augmentation techniques are applied.
\textbf{2)} In Stage II, a hyperbolic encoder-decoder projects visual features from Stage I into hyperbolic space using supervised classification loss and then reconstructs them in Euclidean space. This process captures hierarchical relationships within the image corpus, facilitating the generation of diverse images by manipulating latent codes within the same category.

We design a two-stage model for two reasons:
\textbf{1)} jointly and end-to-end optimizing the whole pipeline is challenging for satisfactory results as multiple loss functions must be optimized for various image generation functionalities; \textbf{2)} end-to-end training requires extensive labeled data, which is difficult to obtain. By staging the pipeline, we can leverage existing datasets collected for general vision tasks and avoid the need for new task-specific paired data.


\begin{figure*}[t]
  \centering
  \setlength{\abovecaptionskip}{0.1cm}   
  \setlength{\belowcaptionskip}{-0.5cm}   
   \includegraphics[width=0.7\linewidth]{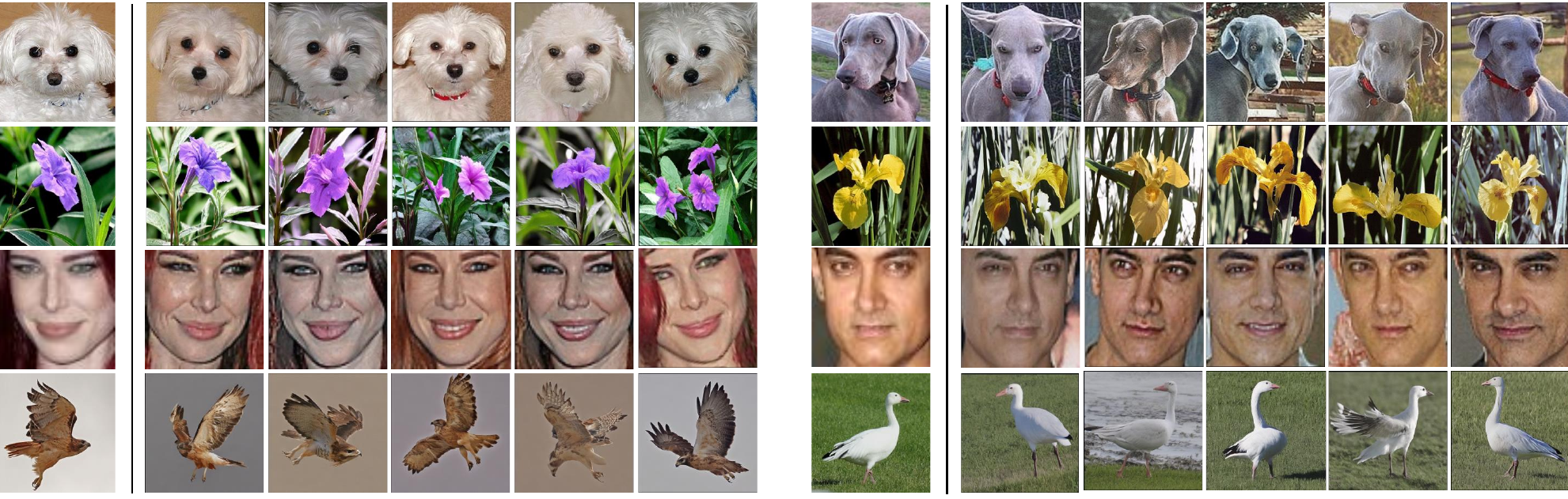}

   \caption{\textbf{One-shot image generation from \modelname{}} on Animal Faces, Flowers, VGGFaces, and NABirds.}
   \label{fig: generation}
\end{figure*}

\vspace{0.05cm}
\noindent\textbf{Diffusion Autoencoders.} 
\label{subsec: Diffusion Autoencoders}
In this stage, we aim to generate new images $x'$ of the reference image $x$ with rich stochastic details (\eg, hair, fur, color, \etc) while maintaining the identity and category of the image unchanged.

To address the limitations of GAN-based methods, we adopt the approach of DiffAE~\cite{preechakul21diffae}, leveraging a pre-trained SD model. 
The SD \textbf{decoder}, $\epsilon_\theta\left(\boldsymbol{z}_t, t, \boldsymbol{c}\right)$, predicts noise based on timestep $t$ and a high-level semantic condition $\boldsymbol{c}$, which is learned by a \textbf{semantic encoder} mapping input image $x$ to a meaningful latent code $\boldsymbol{c}$. SD also functions as a \textbf{stochastic encoder} that captures a low-level “stochastic” subcode $\boldsymbol{z}_T$, as illustrated in \cref{fig: pipeline}.

For the original SD models, the condition $\boldsymbol{c}$ is the given text and is usually processed by a pre-trained CLIP~\cite{Radford21CLIP} text encoder, outputting 77 tokens. Hence, a naive solution is to directly replace it with CLIP image embeddings to capture high-level semantics. However, this naive solution makes the model easy to remember instead of understanding the context information and copying the content, arriving at a trivial solution. We apply two tricks to avoid this: \textbf{1) Strong data augmentations} $\mathcal{A}$ (\eg, flip, rotation, blur, and elastic transform) on the reference image $x$ to break down the connection with the source image and encourage identity-invariant feature learning; \textbf{2) Content Bottleneck.} To promote diversity, we use only the CLIP image encoder’s class token. It compresses the reference image from spatial size $224 \times 224 \times 3$ to a vector of dimension $1 \times 1024$ for a compact representation, aligning it with the original CLIP text feature space of SD via additional fully connected layers (MLP) that inject features into the diffusion process through cross-attention. This mechanism forces the semantic encoder to focus on the main object while ignoring backgrounds and identity-irrelevant features. Thus, given reference image $x$, the condition is defined as: $\boldsymbol{c}=\operatorname{MLP}\left(\operatorname{CLIP}\left(\mathcal{A}\left(x\right)\right)\right)$.

In this setup, noise is progressively added to $\boldsymbol{z}_0$ ($x$ encoded by VAE) to produce a noisy latent $\boldsymbol{z}_{t}$, where $t$ denotes the number of noise additions. To utilize the strong image prior learned by SD and CLIP, only the MLP is trainable during the training process. The UNet $\epsilon_\theta$ learns to predict the added noise with the condition $\boldsymbol{c}$:
\begin{equation}
\label{eq: diffusion loss}
\left.\mathcal{L}_{align}=\mathbb{E}_{\boldsymbol{z}_0, t, \boldsymbol{c}, \epsilon \sim \mathcal{N}(0,1)}\left[\| \epsilon-\epsilon_\theta\left(\boldsymbol{z}_t, t, \boldsymbol{c}\right)\right) \|_2^2\right].
\end{equation}

Besides decoding, the SD model can also be used to encode an input image latent $\boldsymbol{z}_0$ to the stochastic subcode $\boldsymbol{z}_T$ by running DDIM sampling process backward~\cite{Song21DDIM}:
\begin{equation}
    \boldsymbol{z}_{t+1}=\sqrt{\frac{\alpha_{t+1}}{\alpha_t}} \boldsymbol{z}_t+\left(\sqrt{\frac{1}{\alpha_{t+1}}-1}-\sqrt{\frac{1}{\alpha_t}-1}\right) \cdot \epsilon_\theta\left(\boldsymbol{z}_t, t, \boldsymbol{c}\right).
\end{equation}
For the definition of $\alpha_t$ and additional details, please refer
to Sec. B of the supplementary material (SM).

This process functions as a stochastic encoder, where $\boldsymbol{z}_T$ retains details not captured by the limited-capacity semantic representation $\boldsymbol{c}$. By integrating both semantic and stochastic encoders, the diffusion autoencoder preserves full image details while providing a high-level representation $\boldsymbol{c}$ for downstream tasks. Notably, the stochastic encoder is omitted during training (\cref{eq: diffusion loss}) and applied post-training to compute $\boldsymbol{z}_T$, enabling control over image diversity.


\vspace{0.05cm}
\noindent\textbf{Hyperbolic Image Encoder.} 
\label{subsec: hyperbolic embedding}
Despite the category-invariant feature extraction in Stage I, the generated image variations remain limited. Thus, Stage II aims to capture hierarchical relationships within the image, enabling latent code editing to enhance diversity while preserving object identity.


To manipulate latent code in hyperbolic space, we need to define a bijective map from $\mathbb{R}^n$ to $\mathbb{D}_c^n$ to map Euclidean vectors to the hyperbolic space and vice versa. A manifold is a differentiable topological space that locally resembles the Euclidean space $\mathbb{R}^n$~\cite{Lee06,Lee13}. For $x \in \mathbb{D}^n$, one can define the tangent space $T_x \mathbb{D}^n_c$ of $\mathbb{D}^n_c$ at $x$ as the first-order linear approximation of $\mathbb{D}^n_c$ around $x$. Therefore, this bijective map can be performed by exponential and logarithmic maps.
Specifically, the \textit{exponential map} $\exp _{\mathbf{x}}^c: T_{\mathbf{x}} \mathbb{D}_c^n \cong \mathbb{R}^n \to \mathbb{D}_c^n$, maps from the tangent spaces into the manifold. While the \textit{logarithmic map} $\log _{\mathbf{x}}^c: \mathbb{D}_c^n \to T_{\mathbf{x}} \mathbb{D}_c^n \cong \mathbb{R}^n$ is the reverse map of the exponential map.

We train a hyperbolic image encoder to map latent codes from Euclidean to hyperbolic space using exponential and logarithmic maps at the origin $\mathbf{0}$. Given a latent vector $\boldsymbol{c} \in \mathbb{R}^{1 \times 1024}$ from Stage I in CLIP-space, a 5-layer single-head Transformer encoder $\operatorname{E}$ reduces its dimensionality to $\mathbb{R}^{1 \times 512}$ before mapping it to hyperbolic space via the exponential map. A hyperbolic feed-forward layer~\cite{Ganea18} then produces the hierarchical representation $\boldsymbol{c}_h$, formulated as: 
    $\boldsymbol{c}_{h} = f^{\otimes_c}(\exp_{\mathbf{0}}^c(\operatorname{E}(\boldsymbol{c}))),$
where $f^{\otimes_c}$ denotes the Möbius linear layer mapping Euclidean to hyperbolic space. To project $\boldsymbol{c}_h$ back to the CLIP image space and reconstruct the latent code $\boldsymbol{c}'$, we apply a logarithmic map followed by a 30-layer single-head Transformer decoder $\operatorname{D}$:
$\boldsymbol{c}' = \operatorname{D}(\log _{\mathbf{0}}^c(\boldsymbol{c}_{h})),$
which is then fed into the cross-attention layer of the SD model to reconstruct the image $x'$.

\noindent\textbf{Loss functions.} To learn a semantic hierarchy in hyperbolic space, we minimize the distance between latent codes of similar images while increasing the distance between dissimilar ones. For multi-class classification on the Poincaré disk (\cref{sec: Hierarchical Learning}), we extend multinomial logistic regression (MLR)~\cite{Ganea18} to hyperbolic space by incorporating a linear layer and computing the \textbf{Hyperbolic Loss} via the \textit{negative log-likelihood} (NLL):
\begin{equation}
\label{eq:NLLLoss}
    \mathcal{L}_{\text {hyper}} = -\frac{1}{N}\sum_{n=1}^{N}\log(p_n),
\end{equation}
where $N$ is the batch size and $p_n$ represents the predicted probability of the correct class.

As discussed in \cref{sec: Hierarchical Learning}, distances in the Poincaré disk grow exponentially with radius. To minimize \cref{eq:NLLLoss}, fine-grained image embeddings are pushed toward the disk's boundary to maximize inter-category distances, while abstract image embeddings—representing shared features across categories—remain near the center.

 To ensure that the network projects back to the CLIP image space, we use L-2 distance \textbf{reconstruction loss} and \textbf{cosine similarity loss} to reconstruct $\boldsymbol{c}'$ as the same as the original $\boldsymbol{c}$ in CLIP image space:
\begin{equation}
\label{eq: reconstruction loss}
    \mathcal{L}_{\text{rec}}(\boldsymbol{c}, \boldsymbol{c}')=\|\boldsymbol{c}-\boldsymbol{c}'\|_2+1-\cos \left(\boldsymbol{c}, \boldsymbol{c}'\right).
\end{equation}
The \textbf{overall loss function} is:
\begin{equation}
\label{eq: overall loss}
    \mathcal{L} = \mathcal{L}_{\text {hyper}} + \lambda \cdot \mathcal{L}_{\text{rec}},
\end{equation}
where $\lambda$ is a trade-off adaptive parameter. This curated set of loss functions ensures the model learns the hierarchical representation and reconstructs images. More details can be found in Sec. A of the supplementary material.

\subsection{Pseudo-Labeling}
\label{sec: Pseudo-Labeling}
As discussed in \cref{sec:intro}, a key limitation of previous few-shot image generation methods is the reliance on labeled data, which is also evident in our earlier approach (\ie, \cref{eq:NLLLoss}). To mitigate this, we leverage a pre-trained CLIP model to predict the pseudo-label. Given an image and a predefined set of class names (\eg, Golden Retriever,  Spaniel), we utilize CLIP to compute semantic similarities between the image embedding and class embeddings. The pseudo-label is determined by selecting the class with the highest similarity score, which represents the closest semantic match to the image. This process enables label-free classification by leveraging CLIP's cross-modal understanding capabilities. See the Sec. A in supplementary material for further details.

\subsection{Hyperbolic Latent Editing}
\label{sec: hyperbolic latent editing}
To generate diverse images with varying identity-irrelevant features, we edit hyperbolic latent codes via interpolation between two images or by applying random perturbations. In hyperbolic space, the shortest path between two points is defined by the geodesic under the induced distance (\cref{eq3-distance}). The geodesic between two embeddings, $\boldsymbol{c}_{hi}$ and $\boldsymbol{c}_{hj}$, is denoted as $\gamma_{\boldsymbol{c}_{hi} \rightarrow \boldsymbol{c}_{hj}}(t)$. For perturbation-based generation, we first rescale the embedding $\boldsymbol{c}_h$ of a given image $x_i$ to a fixed radius $r_\mathbb{D}$, then sample a random vector $\boldsymbol{c}_{hj}$ from seen category embeddings, also constrained to $r_\mathbb{D}$. The geodesic between them defines the perturbation direction, enabling diverse image generation. More details and formulas are provided in Sec. C of the appendix.

\begin{figure}[t]
  \centering
  \vspace{0.2cm}  
  \setlength{\abovecaptionskip}{0.0cm}   
  \setlength{\belowcaptionskip}{-0.1cm}   
   \includegraphics[width=0.92\linewidth]{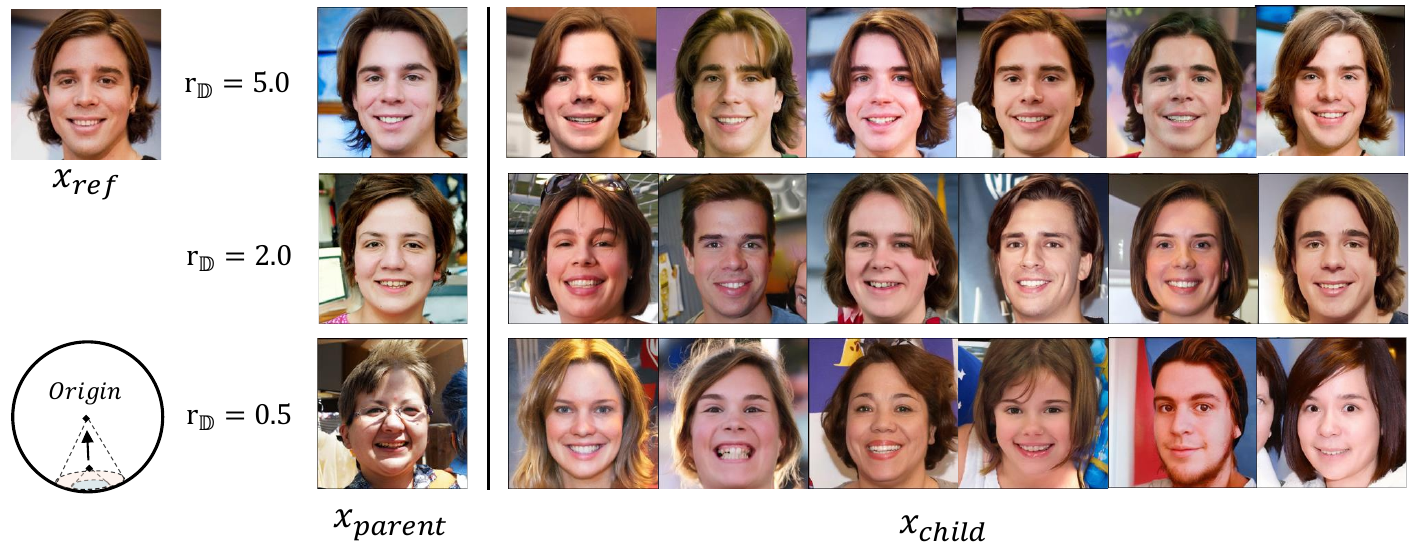}

   \caption{\textbf{Images with hierarchical semantic similarity generated by \modelname{}} by sampling ``child'' images of their ``parent'' images, where $r_\mathbb{D}$ represents the hyperbolic distance from the ``parent'' images to the center of the Poincaré disk. As the reference image $x$ moves from the edge to the center (from fine-grained/certain to abstract/ambiguous), the ``children'' of the reference image become more diverse and less similar to the reference image.}
   \label{fig: Hierarchical_Generation}
\end{figure}

\begin{figure}[t]
  \centering
  \setlength{\abovecaptionskip}{0.1cm}   
  \setlength{\belowcaptionskip}{-0.3cm}   
   \includegraphics[width=0.85\linewidth]{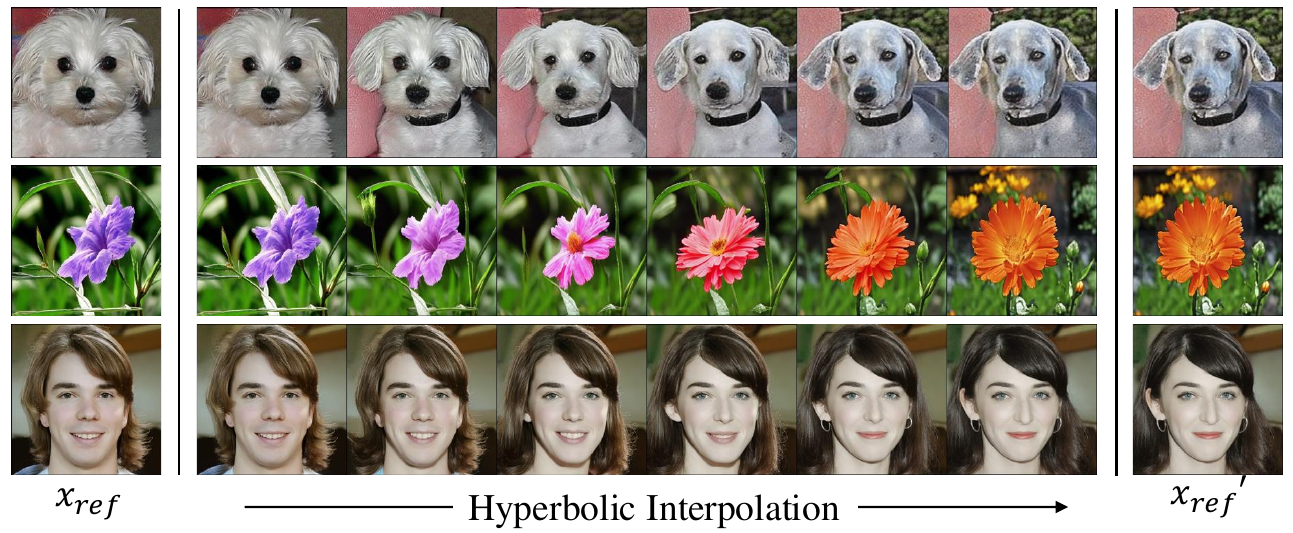}

   \caption{\textbf{Interpolations in hyperbolic space along the edge of the Poincaré disk} (with $r_\mathbb{D} = 6.2126$) on four datasets. Zoom in to see the details.}
   \label{fig: Interpolation}
\end{figure}

\section{Experiment}
\label{sec:experiment}
\subsection{Implementation Details}
We choose Stable Diffusion V2.1~\citep{Rombach22ldm} as the base generator for the base generative model.  
During training, we set the image resolution to $512\times512$.
The dimension of the latent code in hyperbolic space is chosen to be $512$.
More details can be found in the Sec. A of the supplementary material.

\subsection{Datasets}
We evaluate our method on Animal Faces~\cite{Liu19Few}, Flowers~\cite{Nilsback08}, VGGFaces~\cite{Parkhi15}, FFHQ~\cite{Karras19}, and NABirds~\cite{Horn15Nabirds} following the settings described in~\cite{Ding23, Li23HAE}. Due to the low resolution ($64\times 64$) of the VGGfFces dataset, we use FFHQ~\cite{Karras19} to fine-tune the model pre-trained on VGGFaces without supervision and visualize images of human faces with FFHQ.

\begin{figure}[t]
  \centering
  \vspace{-0.1cm}  
  \setlength{\abovecaptionskip}{0.2cm}   
  \setlength{\belowcaptionskip}{0.0cm}   
   \includegraphics[width=0.9\linewidth]{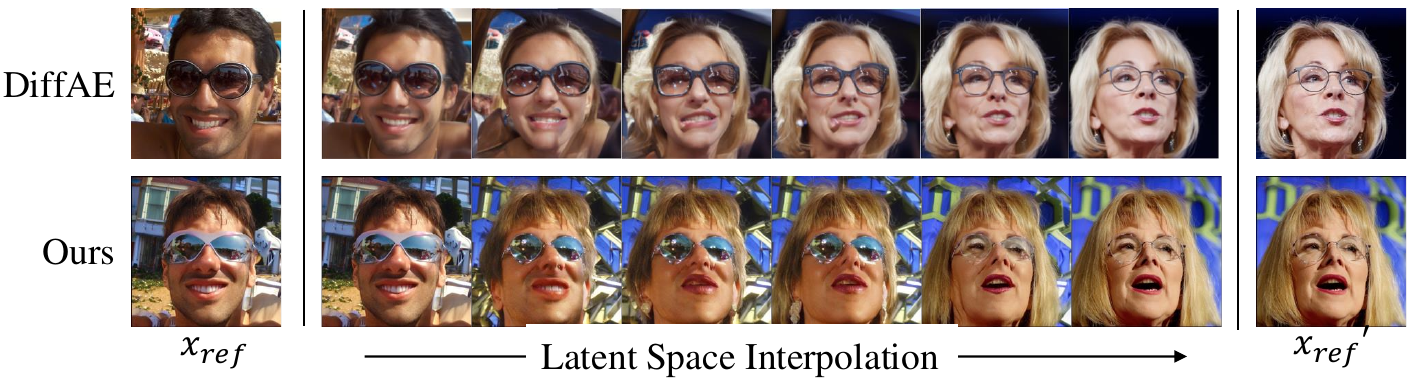}

   \caption{\textbf{Comparison of the latent space interpolations} between DiffAE~\cite{preechakul21diffae} (Euclidean Space) and \modelname{} (Hyperbolic Space).}
   \label{fig: Interpolation_Comparison}
   \vspace{-1.0ex}
\end{figure}

\begin{figure}[t]
  \centering
  \vspace{-0.0cm}  
  \setlength{\abovecaptionskip}{0.2cm}   
  \setlength{\belowcaptionskip}{-0.5cm}   
   \includegraphics[width=0.85\linewidth]{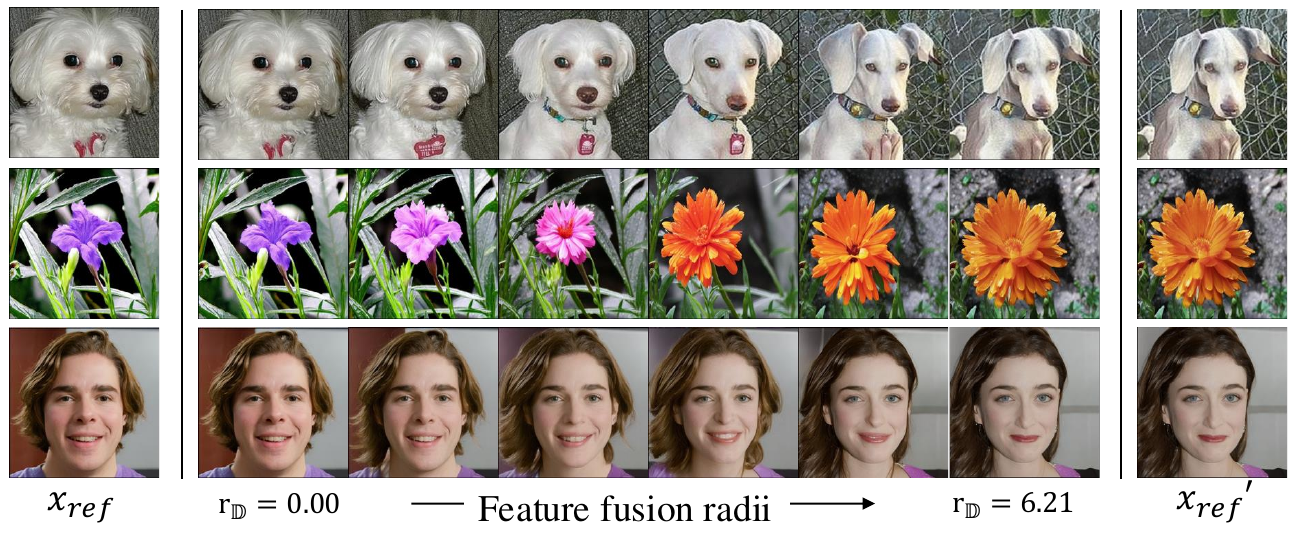}
   \caption{\textbf{Feature fusion from the edge to the center of the Poincaré disk} between two images on four datasets. Zoom in to see the details.}
   
   \label{fig: Feature_Fusion}
   \vspace{-1.0ex}
\end{figure}

\begin{table*}
    \centering
    \resizebox{0.92\linewidth}{!}{
    \setlength{\abovecaptionskip}{0.1cm}   
    \setlength{\belowcaptionskip}{-0.2cm}   
    \begin{tabular}{c|ccccccccc}
        \toprule
        \multicolumn{1}{l}{\multirow{2}{*}{Method}} & \multicolumn{1}{c}{\multirow{2}{*}{Settings}} & \multicolumn{2}{c}{Flowers} & \multicolumn{2}{c}{Animal Faces} & \multicolumn{2}{c}{VGG Faces*} & \multicolumn{2}{c}{NA Birds} \\ 
        & & FID($\downarrow$) &  LPIPS($\uparrow$) & FID($\downarrow$) &  LPIPS($\uparrow$) & FID($\downarrow$) &  LPIPS($\uparrow$) & FID($\downarrow$) &  LPIPS($\uparrow$)\\
        \midrule
        DAWSON~\cite{Liang20} & $3$-shot & 188.96 & 0.0583 &208.68 & 0.0642 & 137.82 & 0.0769 & 181.97 & 0.1105\\
        F2GAN~\cite{Hong20F2} & $3$-shot & 120.48 & 0.2172 &117.74 & 0.1831 & 109.16 & 0.2125 & 126.15 & 0.2015\\
        WaveGAN~\cite{yang22wavegan} & $3$-shot & 42.17 & 0.3868 & 30.35 & 0.5076 & 4.96 & 0.3255 & - & - \\
        F2DGAN~\cite{Zhou24F2DGAN} & $3$-shot & 38.26 & 0.4325 & 25.24 & 0.5463 & \underline{4.25} & 0.3521 & - & - \\
        \midrule
        DeltaGAN~\cite{Hong22Delta} & $1$-shot & 109.78 & 0.3912 &89.81 & 0.4418 & 80.12 & 0.3146 & 96.79 & 0.5069\\
        SAGE~\cite{Ding23} & $1$-shot & 43.52 & 0.4392 & 27.43 & 0.5448 & 34.97 & 0.3232  & 19.45 & 0.5880 \\
        HAE~\cite{Li23HAE} & $1$-shot & 50.10 & 0.4739 & 26.33 & 0.5636 & 35.93 & 0.5636 & 21.85 & 0.6034 \\
        LSO~\cite{zheng23lso} & $1$-shot & 35.87 & 0.4338 & 27.20 & 0.5382 & \textbf{4.15} & 0.3834 & - & - \\
        
        \midrule
        HypDAE (Euc) & $1$-shot &  25.06 &  0.7420 & 20.72 & 0.7288 & 6.52 & 0.5429 & 8.40 & 0.7904 \\
        HypDAE (Real) & $1$-shot & \textbf{23.96} & \underline{0.7595} & \underline{14.31} & \underline{0.7415} & 6.25 & \textbf{0.5685} & \underline{7.64} & \underline{0.7959} \\
        HypDAE (Pseudo) & $1$-shot & \underline{24.43} & \textbf{0.7630} & \textbf{13.14} & \textbf{0.7431} & 5.96 & \underline{0.5560} & \textbf{7.57} & \textbf{0.7966} \\
        \bottomrule
        
    \end{tabular}
    }
    \vspace{-1.0ex}
    \caption{FID($\downarrow$) and LPIPS($\uparrow$) of images generated by different methods for unseen categories on four datasets. \textbf{Bold} indicates the best results and \underline{underline} indicates the second best results. VGGFaces is marked with * because different methods report different numbers of unseen categories on this dataset (\textit{e.g.} 552 in LoFGAN, 96 in DeltaGAN, 497 in L2GAN, and 572 in HypDAE, HAE, and SAGE). We do not compare with diffusion-based models for editing or customization purposes as none of them focus on our task in the same settings.}
    \label{tab: quantitative_comparison}
\end{table*}

\begin{figure*}[t]
  \centering
  \setlength{\abovecaptionskip}{0.1cm}   
  \setlength{\belowcaptionskip}{-0.2cm}   
   \includegraphics[width=0.91\linewidth]{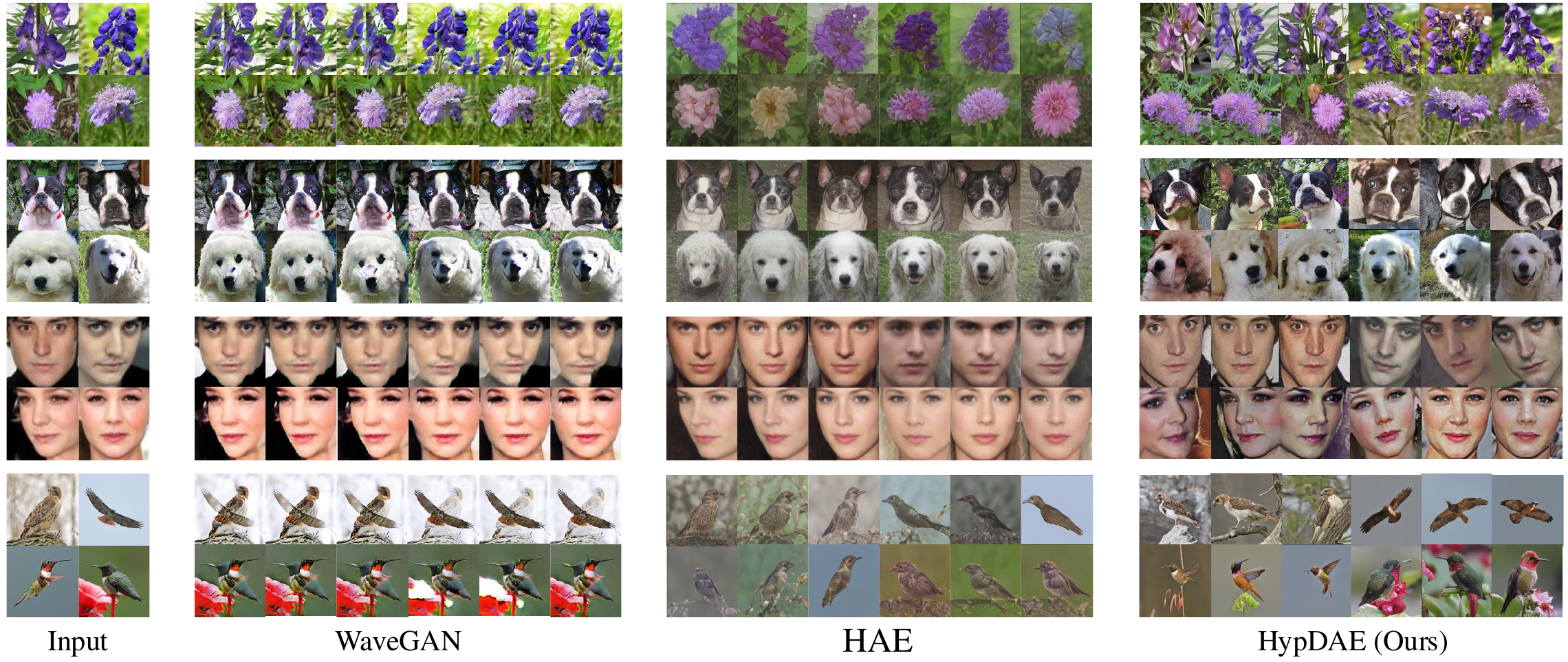}
   \caption{\textbf{Comparison between images generated by WaveGAN, HAE, and \modelname{}} on Flowers, Animal Faces, VGGFaces and NABirds. Note: WaveGAN uses a 2-shot setting; HAE and \modelname{} are both in a 1-shot setting. \textbf{Zoom in to see the details.}}
   \label{fig: Comparison}
   \vspace{-1.0ex}
\end{figure*}

\begin{figure}[t]
  \centering
  \setlength{\abovecaptionskip}{0.2cm}   
  \setlength{\belowcaptionskip}{-0.4cm}   
   \includegraphics[width=0.9\linewidth]{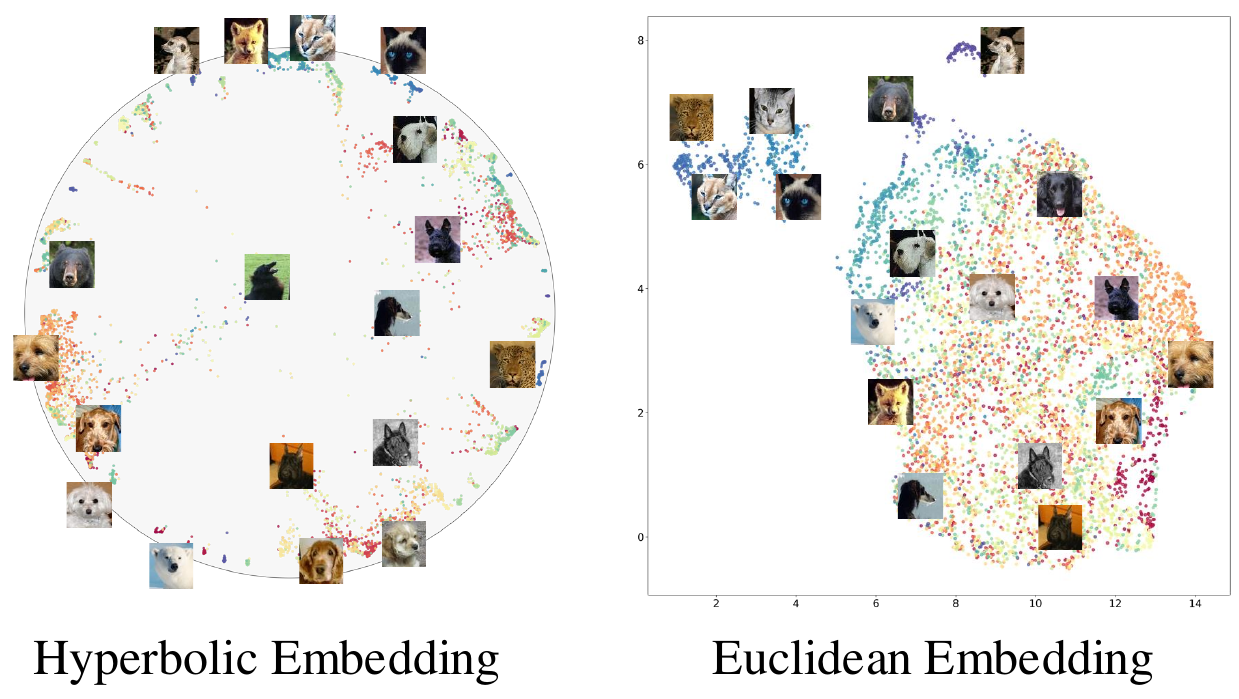}
\vspace{-1.0ex}
   \caption{\textbf{UMAP visualization} of Hyperbolic and Euclidean 2-D embeddings of AnimalFaces dataset. Images with clear identities are clustered and positioned near the boundary, while ambiguous samples are located near the center in hyperbolic space.}
   \label{fig: umap}
\vspace{-1.0ex}
\end{figure}

\subsection{Analysis of Hierarchical Feature Editing} \label{sec: Analysis of Hierarchical Feature Editing} We examine the properties of the hierarchical representations, focusing on how attribute levels correspond to latent code locations in hyperbolic space. As noted in \cref{sec: Hierarchical Learning}, there is a continuum from fine-grained to abstract attributes, corresponding to points from the periphery to the center of the Poincaré disk. We define the hyperbolic radius $r_\mathbb{D}$\footnote{The radius of the Poincaré disk in our experiment is about $6.2126$} as the distance of a latent code to the disk’s center. To investigate the effect of $r_\mathbb{D}$, we conduct experiments with various radii.

\begin{figure}[t]
  \centering
  \vspace{-0.0cm}  
  \setlength{\abovecaptionskip}{0.2cm}   
  \setlength{\belowcaptionskip}{-0.2cm}   
   \includegraphics[width=0.88\linewidth]{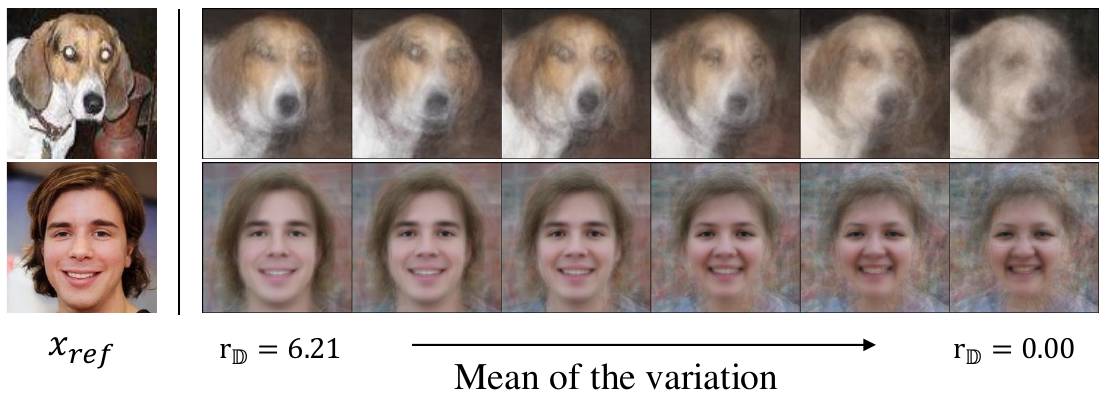}
   \caption{\textbf{The mean of 20 randomly sampled images} by moving the latent codes from the edge to the center of the Poincaré disk.}
   \label{fig: Radius}
   \vspace{-1.0ex}
\end{figure}

\noindent \textbf{Hierarchical Image Sampling and Interpolation.}
\cref{fig: umap} shows 2-D embeddings of Animal Faces in the Poincaré disk, visualized via UMAP~\cite{mcinnes2018umap-software}. By varying the radii of “parent” images, \modelname{} controls the semantic diversity of generated images (\cref{fig: Hierarchical_Generation}). Results indicate that identity-relevant attributes change below $r_\mathbb{D} \approx 2.0$, while identity-irrelevant attributes vary above $r_\mathbb{D} \approx 5.0$, demonstrating that $r_\mathbb{D}$ correlates with attribute levels. As $r_\mathbb{D}$ decreases, identities become more abstract, resulting in greater diversity among “children” images, as seen in \cref{fig: Radius}. Smooth interpolation between attributes is achievable in hyperbolic space (\cref{fig: Interpolation}) without distortion, as shown in \cref{fig: Interpolation_Comparison}, confirming that \modelname{} enables geodesic, hierarchical editing.

\noindent \textbf{Hierarchical Feature Fusion.}
\modelname{} also supports infinite semantic levels for attribute fusion. In \cref{fig: Feature_Fusion}, attributes from two images are combined at varying levels along the radius from the disk’s edge to the center, achieving smooth fusion without the limitations of GAN-based generators, which typically support only finite fusion levels (\eg, 18 levels in StyleGAN2’s $\mathcal{W^+}$-space).

\subsection{Ablation Study}
\label{sec: ablation study}
\begin{figure}[t]
  \centering
  \vspace{-0.0cm}  
  \setlength{\abovecaptionskip}{0.2cm}   
  \setlength{\belowcaptionskip}{-0.2cm}   
   \includegraphics[width=0.88\linewidth]{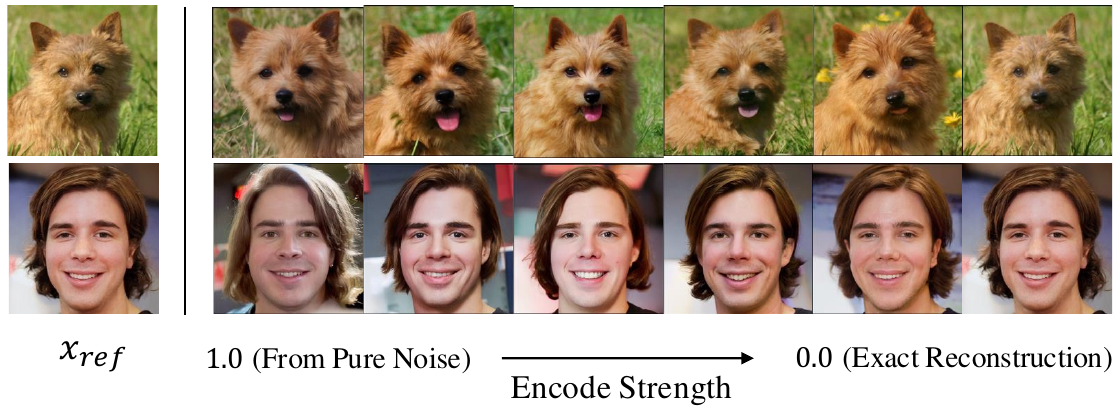}
   \caption{\textbf{Ablation study} on the influence of the encoding strength of the stochastic encoder on four datasets (strength equals $1$ means $x_0$ is fully deconstructed, \ie, $x_T$ is a Gaussian noise).}
   \label{fig: Subcodes}
   \vspace{-1.0ex}
\end{figure}

Three ablation studies assess the impact of the stochastic encoder and hyperbolic embedding radius on few-shot image quality and diversity. As shown in \cref{fig: Subcodes}, the stochastic encoder regulates the similarity between generated and reference images, where stronger encoding enhances similarity at the cost of diversity. \Cref{tab: ablation} further demonstrates that the hyperbolic embedding radius $r_\mathbb{D}$ affects image quality and diversity—smaller radii increase diversity but may alter high-level attributes. The optimal trade-off is observed at $r_\mathbb{D} \approx 5.5$. Additionally, \cref{tab: ablation_radii} reveals that identity shifts toward perturbed images when $r_\mathbb{D}<4.5$, as measured by CLIP-S (CLIP similarity between generated images and reference images) and CLIP-P (CLIP similarity between generated images and perturbed images).

\begin{figure}[t]
  \centering
  \setlength{\abovecaptionskip}{0.2cm}   
  \setlength{\belowcaptionskip}{-0.3cm}   
   \includegraphics[width=0.91\linewidth]{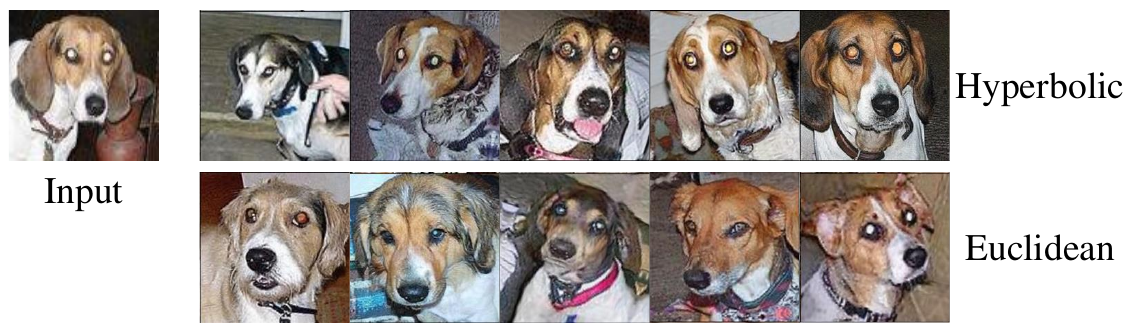}

   \caption{\textbf{Ablation study} of few shot image generation by \modelname{}(Hyp) and \modelname{}(Euc).}
   \label{fig: hyper vs euc}
   \vspace{-1.0ex}
\end{figure}

\noindent\textbf{Euclidean versus Hyperbolic.} To evaluate performance gains, we re-trained \modelname{} (Euc) in Euclidean space using \cref{eq:NLLLoss}. As in \cref{tab: quantitative_comparison} and \cref{fig: hyper vs euc}, the hyperbolic space improves performance due to enhanced latent code disentanglement~\cite{Ge22}. This is further validated by UMAP visualizations in \cref{fig: umap}. More details in Sec. E of the SM.

\noindent\textbf{Pseudo versus Real Label.} The accuracy of the pseudo labels predicted in \cref{sec: Pseudo-Labeling} for the four datasets are: Animal Faces (48.9\%), Flowers (78.6\%), VGG Faces (41.5\%), and NABirds (39.1\%). We re-trained \modelname{} (Pseudo) using the pseudo labels. Results in \cref{tab: quantitative_comparison} indicate that \modelname{} (Pseudo) outperforms \modelname{} (Real) across most benchmarks, suggesting that noise in human-annotated labels may degrade hierarchical representation learning. This finding underscores that \textit{manually labeled data is not essential for few-shot image generation}.

\subsection{Few-Shot Image Generation} As described in \cref{sec: method}, our approach enables few-shot image generation by varying the stochastic subcode $z_T$ or shifting the latent code in a randomly chosen semantic direction within the category cluster. We set $r_\mathbb{D} = 5.5$ and an encoding strength of $0.95$ (\ie, encoding $5\%$ of the information) to achieve this, as illustrated in \cref{fig: generation}. We conduct three experiments demonstrating that \modelname{} achieves promising few-shot generation, with additional examples in the SM.

\begin{table}
\centering

    \resizebox{0.98\linewidth}{!}{
    \setlength{\abovecaptionskip}{0.2cm}   
    \setlength{\belowcaptionskip}{-0.3cm}   
    \begin{tabular}{c|ccccccc}
        \toprule
        Hyp Radius &  6.2 & 6.0 & 5.5 & 5.0 & 4.5 & 4.0 & 3.0\\
        \midrule
        \textbf{FID($\downarrow$)} & 15.18 & 14.67 & \textbf{14.31} & 14.56 & 14.71 & 16.34 & 20.65\\
        \textbf{LPIPS($\uparrow$)} & 0.7035 & 0.7278 & 0.7415 & 0.7643 & 0.7935 & 0.8378 & \textbf{0.8964}\\
        \bottomrule
    \end{tabular}
    }
    \caption{\textbf{Ablation study} of different radii on Animal Faces.}
    \label{tab: ablation}
    \vspace{-1.0ex}
\end{table}

\begin{table}
\centering
\small
    \resizebox{0.98\linewidth}{!}{
    \setlength{\abovecaptionskip}{0.2cm}   
    \setlength{\belowcaptionskip}{-0.3cm}   
    \begin{tabular}{c|ccccccc}
        \toprule
        Hyp Radius & 6.2 & 6.0 & 5.5 & \textbf{5.0} & 4.5 & 4.0 & 3.0 \\
        \midrule
        CLIP-S & 77.37 & 75.41 & 75.15 & 72.16 & 71.45 & 68.20 & 67.89\\
        CLIP-P & 69.62 & 69.89 & 72.35 & 72.86 & 74.25 & 76.34 & 77.00\\
        \bottomrule
    \end{tabular}
    }
    \caption{\textbf{Ablation study} of different radii on Animal Faces.
    }
    
    \label{tab: ablation_radii}
    \vspace{-1.0ex}
\end{table}

\begin{table}
\centering
    \resizebox{0.98\linewidth}{!}{
    \setlength{\abovecaptionskip}{0.2cm}   
    \setlength{\belowcaptionskip}{-0.5cm}   
    \begin{tabular}{c|cccc}
        \toprule
        ~ &  WaveGAN & HAE & SAGE & HypDAE (Ours)\\
        \midrule
        \textbf{Quality}~($\uparrow$) ~ & 1.34&2.67& 2.54 & \textbf{3.45}\\
        \textbf{Fidelity}~($\uparrow$) ~ & 2.05&1.89& 2.68 & \textbf{3.58}\\
        \textbf{Diversity}~($\uparrow$) ~ &1.25&2.53&2.36&\textbf{3.86}\\
        \bottomrule
    \end{tabular}
    }
    \caption{\textbf{User study} on the comparison between our \modelname{} and existing alternatives.
    ``Quality'', ``Fidelity'', and ``Diversity'' measure synthesis quality, object identity preservation, and object diversity.
    Each metric is rated from 1 (worst) to 4 (best).
    }
    \label{table: user-study}
    \vspace{-1.0ex}
\end{table}

\noindent \textbf{Quantitative Comparison with State-of-the-Art.}
To assess fidelity and diversity, we compute FID~\cite{Heusel17} and LPIPS~\cite{Zhang18unreasonable} following the one-shot settings in~\cite{Li23HAE}. Results in \cref{tab: quantitative_comparison} show significant improvements in FID and LPIPS scores on all four datasets except for the FID score on vggfaces due to the low resolution, affirming the advantages of our diffusion-based approach over previous GAN-based methods.

\noindent \textbf{Qualitative Evaluation.}
We compare \modelname{} qualitatively with WaveGAN~\cite{yang22wavegan} and HAE~\cite{Li23HAE}. As shown in \cref{fig: Comparison}, \modelname{} generates highly diverse images with fine-grained fidelity, such as detailed feather textures, which other methods fail to preserve. A user study further evaluates synthesis quality, identity preservation, and category-irrelevant diversity across these methods. As reported in \cref{table: user-study}, \modelname{} consistently outperforms all baselines, with notable improvements in fidelity and diversity, attributed to its effective disentanglement of identity-relevant and irrelevant features. Additional details are provided in Sec. I of the SM.

\section{Conclusion}
\label{sec:conclusion}
In this work, we present \modelname{}, the first diffusion-based method for editing hierarchical attributes in hyperbolic space. By learning the semantic hierarchy of images, our diffusion autoencoder enables continuous and flexible editing of hierarchical features. Experimental results show that \modelname{} significantly outperforms existing GAN-based approaches without human-annotated labels, excelling in both random and customized few-shot generation. Furthermore, it facilitates hierarchical image generation with varying semantic similarity, advancing the state of the art in few-shot image generation. We hope this work inspires future research in this direction.
\\
\textbf{Acknowledgement.} This work is supported by the Shun Hing Institute of Advanced Engineering (SHIAE) Fund (No. 8115074).

{
    \small
    \bibliographystyle{ieeenat_fullname}
    \bibliography{main}
}
\section*{ \huge Supplementary Material\vspace{1em}}
\label{appendix: supplementary material}
\section*{Overview}
This appendix is organized as follows:

\cref{appendix: Implementation Details and Analysis} gives more implementation details of \modelname{}. Sec 3.2 \& Sec 4.1

\cref{appdendix: diffusion_background} gives detailed explanation of diffusion models.

\cref{appendix: Hyperbolic Neural Networks} provides the mathematical formulae used in hyperbolic neural networks. Sec 3.2 \& Sec 3.3

\cref{appendix: Ablation Study} shows more results of the ablation study of \modelname{}. Sec 4.3


\cref{appendix: Comparison with Euclidean space} shows more comparisons between the latent manipulation in hyperbolic and Euclidean space. Sec 4.3

\cref{appendix: Out-of-distribution Few-shot Image Generation} provides examples to show the exceptional out-of-distribution few-shot image generation ability. Sec  4.4

\cref{appendix: Hierarchical Image Generation} shows the images generated with different radii in the Poincaré disk. Sec 4.3

\cref{appendix: Comparison with State-of-the-art few-shot image generation Method} compares the images generated by state-of-the-art few-shot image generation method, \ie WaveGAN~\cite{yang22wavegan}, HAE~\cite{Li23HAE} and our methods \modelname{}. Sec 4.4

\cref{appendix: User Study} gives more details of the user study we conducted. Sec 4.4

\cref{appendix: Additional Examples Generated by HypDAE} gives more examples generated by \modelname{}. Sec 4.4

\section{Implementation Details and Analysis}
\label{appendix: Implementation Details and Analysis}
\noindent\textbf{Stage I. } As mentioned in Sec 3.2, this stage does not require class labels for the images. To promote diversity, we use only the CLIP image encoder’s class token (dimension $1 \times 1024$) for a compact representation, aligning it with the CLIP text feature space via a 5-layer fully connected MLP following the same settings in~\cite{Yang23PBE} that inject features into the diffusion process through cross-attention to replace the text feature in the original stable diffusion model.

We choose Stable Diffusion V2.1~\citep{Rombach22ldm} as the base generator for the base generative model. We set the image resolution to $512\times512$. We choose the Adam optimizer and set the learning rate as $1\mathrm{e}{-5}$.
During the training process, the pre-trained CLIP image encoder and SD V2.1 models are frozen, only the Transformer block for aligning features is trainable. Since the SD model is loaded during the training process, we use 2 $\times$ NVIDIA A800 (80GB) GPUs for training, and the batch size is selected as $24$ for each GPU. We train about $1\mathrm{e}{5}$ steps to get the model to converge on each dataset.

\noindent\textbf{Stage II. } This stage is the only stage that requires the class labels for given images to learn the hierarchical representation. Although class labels are required in Stage II, the model only needs a small number of labeled data for pre-training and pseudo labels can be predicted by CLIP as shown in \cref{appendix: pseudo labeling}. Furthermore, we show exceptional out-of-distribution generation ability in~\cref{appendix: Out-of-distribution Few-shot Image Generation}. For the hyperbolic encoder mentioned in Sec 3.2, we use a single-head 5-layer Transformer block to reduce the dimensionality of the Euclidean latent vector $\boldsymbol{c}$ from $1 \times 1024$ to $1 \times 512$, which is then mapped to hyperbolic space via an exponential map. A hyperbolic feed-forward layer~\cite{Ganea18} produces the final hierarchical representation $z_{\mathbb{D}}$: 
\begin{equation}
    \boldsymbol{c}_{h} = f^{\otimes_c}(\exp_{\mathbf{0}}^c(\operatorname{E}(\boldsymbol{c}))),
\end{equation}
where $\operatorname{E}$ is the Transformer encoder and $f^{\otimes_c}$ is the Möbius translation of feed-forward layer $f$ as the map from Euclidean space to hyperbolic space, denoted as \textit{Möbius linear layer}.  In order to perform multi-class classification on the Poincaré disk defined in Sec 3.1, one needs to generalize multinomial logistic regression (MLR) to the Poincaré disk defined in~\cite{Ganea18}. An extra linear layer needs to be trained for the classification, and the details on how to compute softmax probability in hyperbolic space are shown in \cref{appendix: Hyperbolic Neural Networks}. As mentioned in Sec 3.1, the distance between points grows exponentially with their radius in the Poincaré disk. In order to minimize Eq. (5) in the main paper, the latent codes of fine-grained images will be pushed to the edge of the ball to maximize the distances between different categories while the embedding of abstract images (images have common features from many categories) will be located near the center of the ball. 
Since hyperbolic space is continuous and differentiable, we are able to optimize Eq. (5) with stochastic gradient descent, which learns the hierarchy of the images. 

Then we train a Transformer decoder to project the hyperbolic latent code back to the CLIP image space with exact reconstruction. In practice, this is achieved by firstly applying a logarithmic map followed by a Transformer decoder $\operatorname{D}$:
\begin{equation}
    \boldsymbol{c}' = \operatorname{D}(\log _{\mathbf{0}}^c(\boldsymbol{c}_{h})).
\end{equation}
and $\boldsymbol{c}'$ will be fed into the cross-attention layer of the stable diffusion model to reconstruct the image $x'$. We use a single-head 30-layer Transformer block as the Transformer decoder for Animal Faces~\cite{Liu19Few}, VGGFaces~\cite{Parkhi15}, FFHQ~\cite{Karras19}, and NABirds~\cite{Horn15Nabirds} since these datasets are relatively large. Therefore, a deeper network is needed to reconstruct the latent representation of these large datasets. However, for the Flowers dataset~\cite{Nilsback08}, the number of images is less than $10$ thousand, which is not enough to train a deep neural network. As a consequence, we use a single-head 5-layer Transformer block as the Transformer decoder for Flowers which works well.  

\noindent\textbf{Fine-tuning on FFHQ. } As we mentioned in Sec 4.2 in the main paper, we learn the hierarchy of human faces by training Stage II with VGGFaces first. However, we visualize the human faces with the FFHQ dataset. Note that the FFHQ dataset has no class labels. Therefore, we first use the VGGFaces dataset to learn a good prior of hierarchy among human faces images with supervision, then fine-tune the model with the reconstruction loss $\mathcal{L}_{\text{rec}}$ \textit{only} to teach the model how to reconstruct images with high resolution but maintaining the hierarchical representation prior. The results show the great potential of our model to be fine-tuned on large-scale dataset without supervision.

In Stage II, only the CLIP image encoder is loaded during the training process. Besides, the CLIP image encoder is frozen, and only the lightweight Transformer encoder and decoder are trainable. We use 1 $\times$ NVIDIA RTX 4090 (24GB) GPU for training, and the batch size is selected as $256$. The $\lambda$ in Eq. (7) in the main paper is selected as $0.1$. We choose the AdamW~\cite{Kingma14adam} optimizer and set the learning rate as $1\mathrm{e}{-3}$. A linear learning rate scheduler is used with a step size equal to $5000$, with a multiplier $\gamma=0.5$. We train about $1\mathrm{e}{5}$ steps to get the model to converge on each dataset. 

In addition, as a remark, we choose the largest radius as $6$ in most of our experiments as in hyperbolic space since any vector asymptotically lying on the surface unit $N$-sphere will have a hyperbolic length of approximately $r = 6.2126$, which can be directly calculated by Eq. (2).
 
Although training our model requires considerable computing resources as mentioned before, the runtime cost and resources required for the inference stage are affordable. Our model can inference on a single NVIDIA RTX 4090 GPU (24GB) thanks to our multi-stage training/inference since one does not need to load all models simultaneously.

\noindent\textbf{Pseudo-Labeling. } 
\label{appendix: pseudo labeling}
For Flowers, Animal Faces and NABirds, we utilize the CLIP ViT-B/32 model. Given an image ($x$), we extract its embedding using the CLIP image encoder. Similarly, we compute embeddings for a predefined set of class names ($y_i$) using the CLIP text encoder. The cosine similarity between the image embedding and each class embedding is computed as:
\begin{equation}
\operatorname{sim}(x, y_i) = \frac{f(x) \cdot g(y_i)}{|f(x)| |g(y_i)|},
\end{equation}
where ($f(\cdot)$) and ($g(\cdot)$) denote the CLIP image and text encoders, respectively. The softmax function is applied to convert similarity scores into probabilities. The pseudo-label ($y$) is assigned as the class with the highest probability:
\begin{equation}
y = \underset{j}{\mathrm{argmax}}\,\frac{\exp(\operatorname{sim}(x, y_i))}{\sum_{j} \exp(\operatorname{sim}(x, y_j))}.
\end{equation}
For the VGGFaces dataset, we employ the DeepFace framework with the VGG-Face architecture. DeepFace predicts pseudo-labels by comparing the face embedding of an image with embeddings of known identities in the dataset. Specifically, we construct a reference database by selecting one image per class, and each input image is assigned to the class with the highest similarity score. This approach ensures robust pseudo-labeling by leveraging DeepFace's face recognition capabilities.

\noindent\textbf{Dataset Settings}
We evaluate our method on Animal Faces~\cite{Liu19Few}, Flowers~\cite{Nilsback08}, VGGFaces~\cite{Parkhi15}, FFHQ~\cite{Karras19}, and NABirds~\cite{Horn15Nabirds} following the settings described in~\cite{Ding23, Li23HAE}. 

\noindent \textbf{Animal Faces}. We randomly select 119 categories as seen for training and leave 30 as unseen categories for evaluation.

\noindent \textbf{Flowers}. The Flowers~\cite{Nilsback08} dataset is split into 85 seen categories for training and 17 unseen categories for evaluation.

\noindent \textbf{VGGFaces}. For VGGFaces~\cite{Parkhi15}, we randomly select 1802 categories for training and 572 for evaluation.

\noindent \textbf{NABirds}. For NABirds~\cite{Horn15Nabirds}, 444 categories are selected for training and 111 for evaluation.

\noindent \textbf{FFHQ}. Due to the low resolution ($64\times 64$) of the VGGfFces dataset, we use FFHQ~\cite{Karras19} to fine-tune the model pre-trained on VGGFaces without supervision and visualize images of human faces with FFHQ.

\section{Additional Background - Diffusion Models}\label{appdendix: diffusion_background}

Diffusion Denoising Probabilistic Models (DDPM)~\cite{Ho20DDPM} are generative latent variable models that aim to model a distribution $p_\theta(x_0)$ that approximates the data distribution $q(x_0)$ and easy to sample from. DDPMs model a ``forward process'' in the space of $x_0$ from data to noise. This is called ``forward'' due to its procedure progressing from $x_0$ to $x_T$. Note that this process is a Markov chain starting from $x_0$, where we gradually add noise to the data to generate the latent variables $x_1,\ldots,x_T\in X$. The sequence of latent variables, therefore, follows $q(x_1,\ldots,x_t\mid x_0)=\prod_{i=1}^{t}q(x_t\mid x_{t-1})$, where a step in the forward process is defined as a Gaussian transition $q(x_t\mid x_{t-1}):=N(x_t;\sqrt{1-\beta_t}x_{t-1},\beta_t I)$ parameterized by a schedule $\beta_0,\ldots,\beta_T\in (0,1)$. When $T$ is large enough, the last noise vector $x_T$ nearly follows an isotropic Gaussian distribution.

An interesting property of the forward process is that one can express the latent variable $x_t$ directly as the following linear combination of noise and $x_0$ without sampling intermediate latent vectors: \\[-8pt]
\begin{equation}
  x_t = \sqrt{\alpha_t}x_0+\sqrt{1-\alpha_t}w,~~w\sim N(0,I),\label{eq:xtsamplefromx0}  
\end{equation} \\[-10pt]
where $\alpha_t:=\prod_{i=1}^{t}(1-\beta_i)$.

To sample from the distribution $q(x_0)$, we define the dual ``reverse process'' $p(x_{t-1}\mid x_t)$ from isotropic Gaussian noise $x_T$ to data by sampling the posteriors $q(x_{t-1} \mid x_t)$. Since the intractable reverse process $q(x_{t-1} \mid x_t)$ depends on the unknown data distribution $q(x_0)$, we approximate it with a parameterized Gaussian transition network $p_\theta(x_{t-1}\mid x_t):=N(x_{t-1}\mid \mu_\theta(x_t,t),\Sigma_\theta(x_t,t))$. The $\mu_\theta(x_t,t)$ can be replaced~\cite{Ho20DDPM} by predicting the noise $\epsilon_\theta(x_t,t)$ added to $x_0$ using equation~\ref{eq:xtsamplefromx0}.

\section{Hyperbolic Neural Networks}
\label{appendix: Hyperbolic Neural Networks}
For hyperbolic spaces, since the metric is different from Euclidean space, the corresponding calculation operators also differ from Euclidean space. In this section, we start by defining two basic operations: Möbius addition and Möbius scalar multiplication~\cite{Khrulkov21}, given fixed curvature $c$.

For any given vectors $x, y \in \mathbb{H}^n$, the \textit{Möbius addition} is defined by:

\begin{equation}
\label{mobius addition}
x \oplus_c y=\frac{\left(1-2 c\langle x, y\rangle-c\|y\|_2^2\right) x+\left(1+c\|x\|_2^2\right) y}{1-2 c\langle x, y\rangle+c^2\|x\|_2^2\|y\|_2^2},
\end{equation}
where $\| \cdot \| $ denotes the $2$-norm of the vector, and $\langle \cdot, \cdot \rangle$ denotes the Euclidean inner product of the vectors.

Similarly, the \textit{Möbius scalar multiplication} of a scalar $r$ and a given vector $x \in \mathbb{H}^n$ is defined by:
\begin{equation}
r \otimes_c x=\tan _c\left(r \tan _c^{-1}\left(\|x\|_2\right)\right) \frac{x}{\|x\|_2}.
\end{equation}

We also would like to give explicit forms of the exponential map and the logarithmic map which are used in our model to achieve the translation between hyperbolic space and Euclidean space as mentioned in Sec 3.2.

The \textit{exponential map} $\exp _{x}^c: T_{x} \mathbb{D}_c^n \cong \mathbb{R}^n \to \mathbb{D}_c^n$, that maps from the tangent spaces into the manifold, is given by
\begin{equation}
    \exp _{x}^c(v):=x \oplus_c\left(\tanh \left(\sqrt{c} \frac{\lambda_{x}^c\|v\|}{2}\right) \frac{v}{\sqrt{c}\|v\|}\right).
\end{equation}

The \textit{logarithmic map} $\log _{x}^c(y): \mathbb{D}_c^n \to T_{x} \mathbb{D}_c^n \cong \mathbb{R}^n$ is given by
\begin{equation}
    \log _{x}^c(y):=\frac{2}{\sqrt{c} \lambda_{x}^c} \operatorname{arctanh}\left(\sqrt{c}\left\|-x \oplus_c y\right\|\right) \frac{-x \oplus_c y}{\left\|-x \oplus_c y\right\|}.
\end{equation}

We also provide the formula to calculate the softmax probability in hyperbolic space used in Eq. (5) in the main paper:
Given $K$ classes and $k \in\{1, \ldots, K\}, p_k \in \mathbb{D}_c^n, a_k \in T_{p_k} \mathbb{D}_c^n \backslash\{\mathbf{0}\}$ :
\begin{equation}
\begin{aligned}
    p(y=k \mid x) \propto \exp &\Biggl(\frac{\lambda_{p_k}^c\left\|a_k\right\|}{\sqrt{c}} \sinh ^{-1}\\
    &\biggl(\frac{2 \sqrt{c}\left\langle-p_k \oplus_c x, a_k\right\rangle}{\left(1-c\left\|-p_k \oplus_c x\right\|^2\right)\left\|a_k\right\|}\biggl)\Biggl),\\ 
    & \quad \forall x \in \mathbb{D}_c^n,
\end{aligned}
\end{equation}
where $\oplus_c$ denotes the Möbius addition defined in~\cref{mobius addition} with fixed sectional curvature of the space, denoted by $c$. 

\begin{figure}[t]
    \centering
    \includegraphics[width=0.9\linewidth]{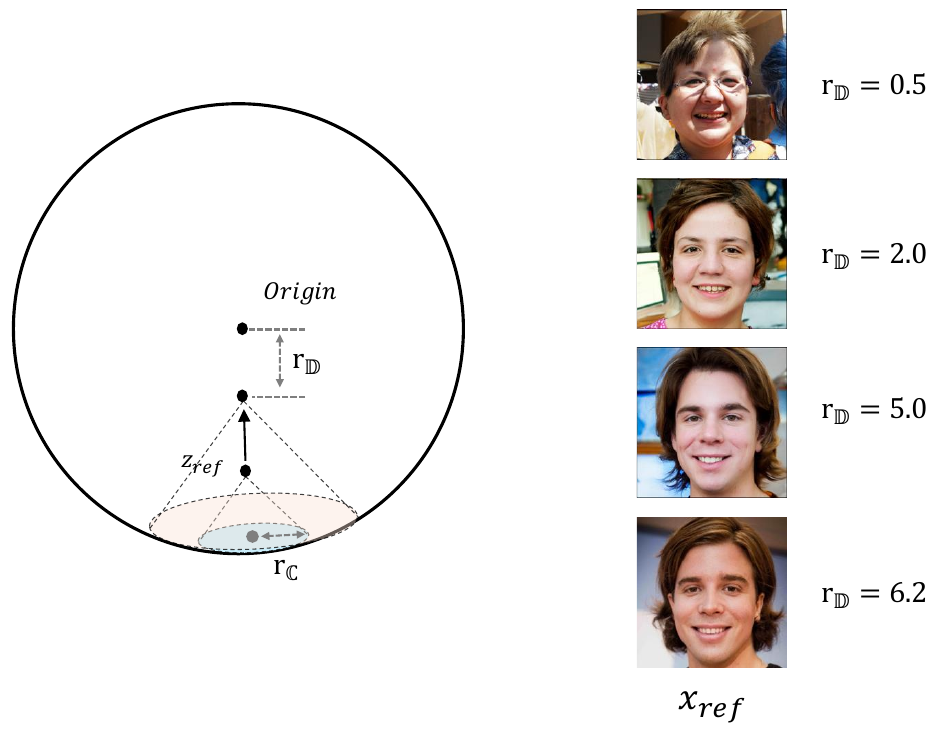}
    \vspace{0ex}
    \caption{\textbf{The illustration of hierarchical data sampling in hyperbolic space}}
    \label{fig: sampling}
\end{figure}

\noindent\textbf{Hierarchical Data Sampling. } As illustrated in~\cref{fig: sampling}, as the latent code $z_{ref}$ of the reference image $x_{ref}$ moves from the edge to the center of the Poincaré disk, the control of the identity of the sampled images becomes weaker and weaker. The identity of the generated images becomes more ambiguous. To conduct hierarchical image sampling in hyperbolic space, one can move the latent code $z_{ref}$ of the reference image $x_{ref}$ towards the origin of the Poincaré disk. Then, sampling latent points among the ``children'' of the rescaled reference images. In practice, the semantic diversity of the generated images can be controlled either by setting different values of $r_\mathbb{D}$, then calculating $r_\mathbb{C}$ based on $r_\mathbb{D}$, or by setting different values of $r_\mathbb{C}$ directly. 

Recall that, in hyperbolic space, the shortest path with the induced distance between two points is given by the geodesic defined in Eq. (2) in the main paper. The geodesic equation between two embeddings $z_{\mathbb{D}\mathbf{i}}$ and $z_{\mathbb{D}\mathbf{j}}$, denoted by $\gamma_{z_{\mathbb{D}i} \rightarrow z_{\mathbb{D}j}}(t)$, is given by 
\begin{equation}
    \gamma_{z_{\mathbb{D}i} \rightarrow z_{\mathbb{D}j}}(t)= z_{\mathbb{D}i} \oplus_c t \otimes_c\left((-z_{\mathbb{D}i}) \oplus_c z_{\mathbb{D}j}\right), \: t \in [0,1],
\end{equation}
where $\oplus_c$ denotes the Möbius addition with aforementioned sectional curvature $c$. Therefore, for hierarchical data sampling, we can first define the value of $r_\mathbb{C}$, then sample random data points in hyperbolic space. If the distance between the sampled data point and $z_{ref}$ is equal to or shorter than the value, we accept the data point. Otherwise, we move the latent codes along the geodesic between the sampled data point and $z_{ref}$ until the distance is within the scope we define. We show more hierarchical image sampling examples in~\cref{appendix: Out-of-distribution Few-shot Image Generation}.

\section{Ablation Study}
\label{appendix: Ablation Study}
There are a few hyperparameters of \modelname{} that control the generation quality and diversity. We conduct ablation studies on each of them in this section.

\noindent\textbf{Hyperbolic Radius. }
By varying the radii of “parent” images, \modelname{} controls the semantic diversity of generated images (\cref{fig: hierarchy_animals}), where the ``parent'' images can be viewed as the image with the average attributes of its children. The quantitative results of the Flowers dataset are presented in~\cref{tab: ablation_radii_supp}. We can see that the diversity increases as the radius becomes smaller, therefore, the value of LPIPS increases accordingly. However, changing too many attributes changes the identity or category of the given images, therefore, the FID decreases when the radius is smaller than $5.5$. In practice, we select $5.5$ as the radius of the parent images for few-shot image generation. 

\begin{figure*}[h]
    \centering
    \includegraphics[width=1.0\textwidth]{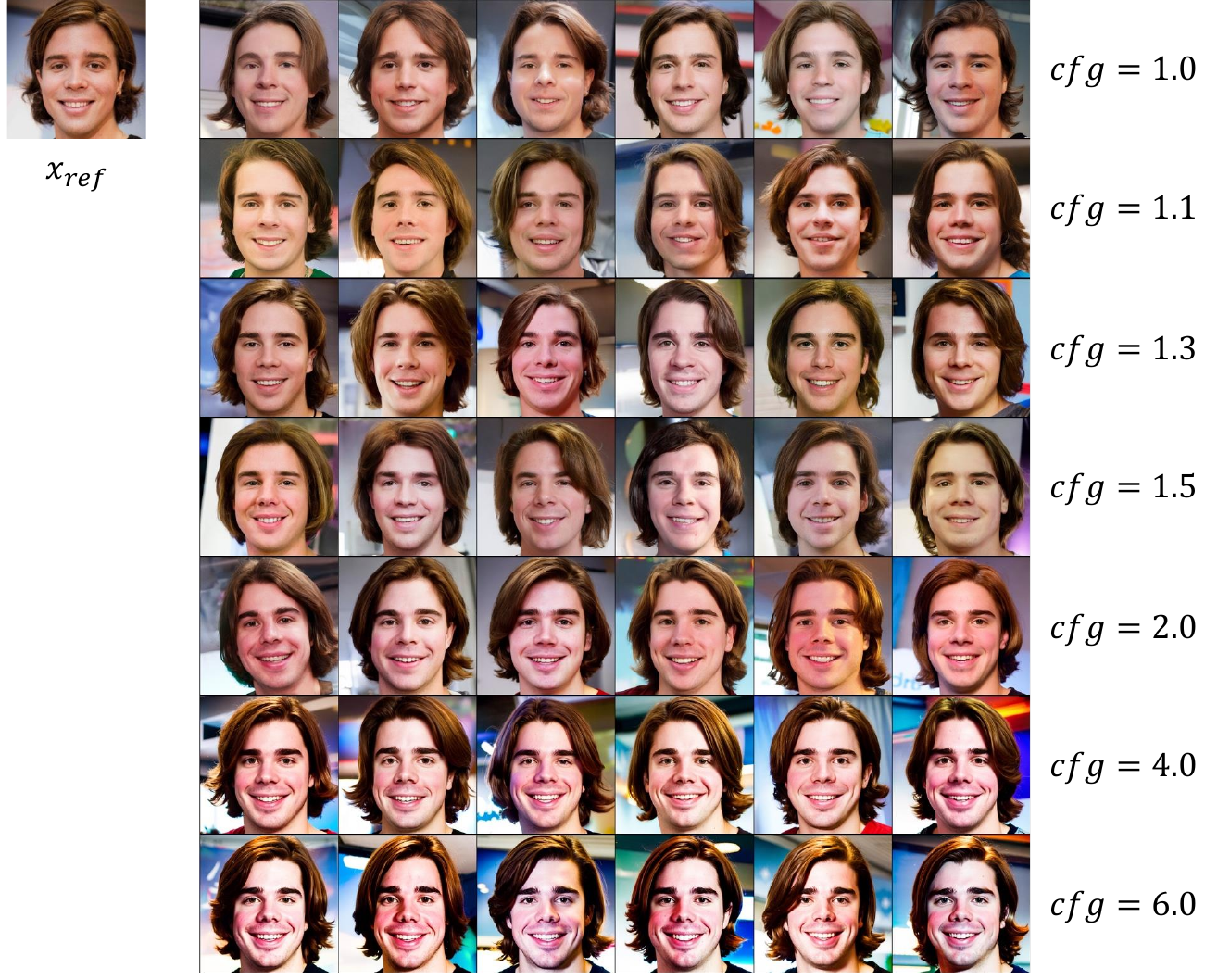}
    \vspace{0ex}
    \caption{\textbf{Ablation study} on the influence of classifier free guidance.}
    \label{fig: cfg}
\end{figure*}

\noindent\textbf{Classifier-free Guidance. }To achieve the trade-off between identity preservation and image harmonization, we find that classifier-free sampling strategy~\citep{Ho22ClassifierFree} is a powerful tool. Previous work~\citep{Tang22Improved} found that the classifier-free guidance is actually the combination of both prior and posterior constraints. In our experiments, we follow the settings in~\citep{Zhang23Controlnet}. 
\begin{equation}
    \epsilon_{\mathrm{prd}}=\epsilon_{\mathrm{uc}}+s\left(\epsilon_{\mathrm{c}}-\epsilon_{\mathrm{uc}}\right), 
\end{equation}

where $\epsilon_{\mathrm{prd}}$, $\epsilon_{\mathrm{uc}}$, $\epsilon_{\mathrm{c}}$, $s$ are the model’s final output, unconditional output, conditional output, and a user-specified weight, respectively. The visualizations are shown in \cref{fig: cfg}, and quantitative results for the Flowers dataset are presented in \cref{tab: ablation_cfg}. Consistent with findings in \cref{tab: ablation_radii_supp}, diversity, measured by LPIPS, increases as the cfg scale grows. However, excessive cfg scaling can alter the identity or category of the input images, leading to a decline in FID when the cfg scale exceeds $1.3$. Based on these results, we select a cfg scale of $1.3$ to achieve optimal few-shot image generation with a balance between fidelity and diversity.

\noindent\textbf{Encoding Strength of the Stochastic Encoder.}
As described in Sec. 3.2, the encoding strength of the stochastic encoder determines the extent of information encoded from the given images. For instance, attributes such as rough posture, color, and style are encoded during the early steps of the diffusion process. A higher encoding strength deconstructs more information from the input images, while a lower encoding strength retains more original information. An encoding strength of $1$ implies full deconstruction, where the initial latent of the denoising process is Gaussian noise. Conversely, an encoding strength of 0 results in exact reconstruction without information loss.

\begin{figure*}[h]
    \centering
    \includegraphics[width=1.0\textwidth]{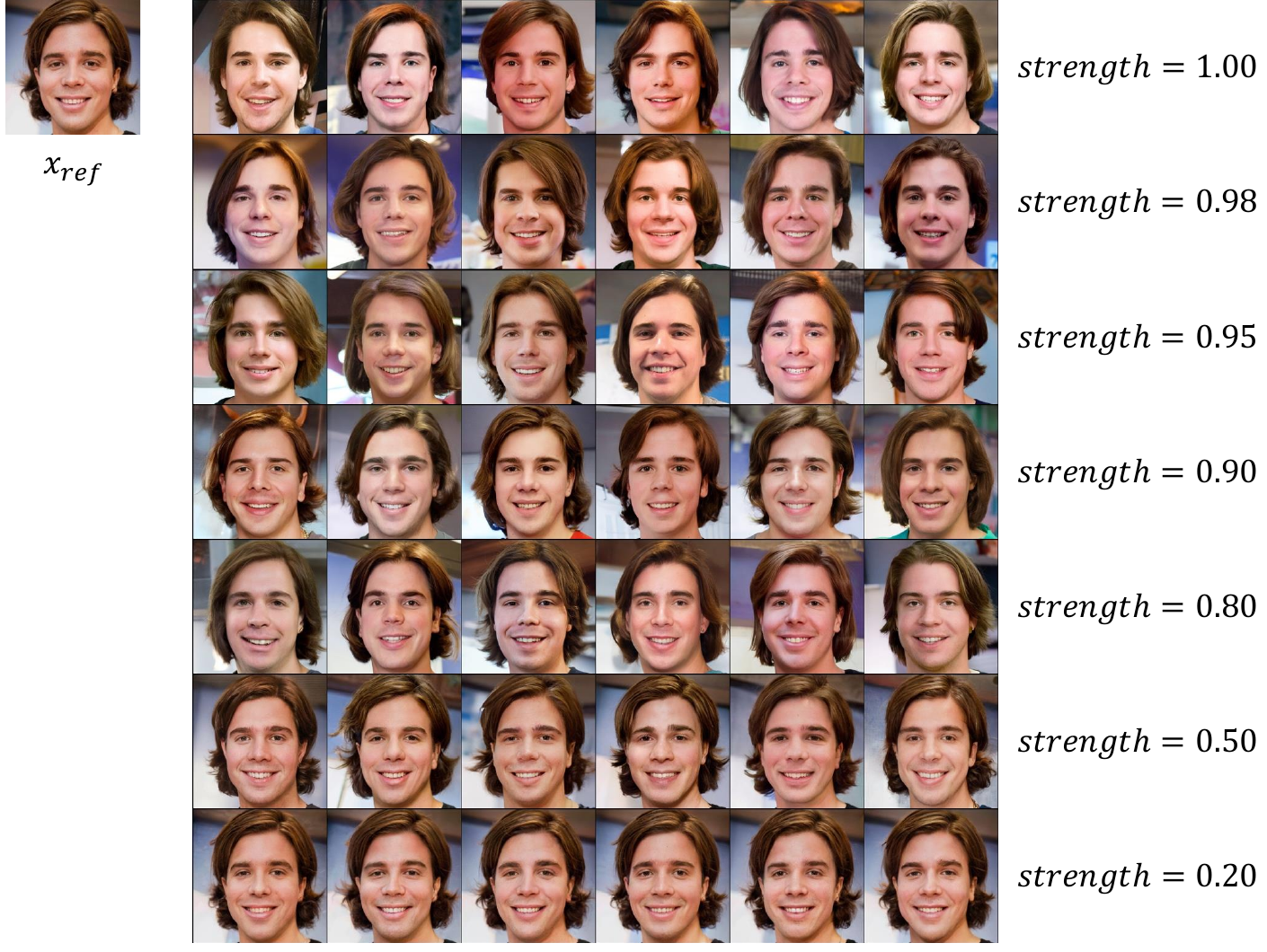}
    \vspace{0ex}
    \caption{\textbf{Ablation study} on the influence of the encoding strength of the stochastic encoder on FFHQ (strength equals $1$ means $x_0$ is fully deconstructed, \ie, $x_T$ is a Gaussian noise).}
    \label{fig: strength}
\end{figure*}

While lower encoding strength preserves the identity and style of the input images, it reduces diversity. This trade-off is visualized in \cref{fig: strength}, and quantitative results for the Flowers dataset are presented in \cref{tab: ablation_strength}. Consistent with \cref{tab: ablation_radii_supp}, diversity, measured by LPIPS, increases with higher encoding strength, while excessive encoding strength can cause changes in identity or category. This is reflected in a decrease in FID when encoding strength exceeds $0.95$ (\ie, $5\%$ of the information is encoded). Based on these findings, we set the encoding strength of the stochastic encoder to $0.95$ to achieve reliable few-shot image generation.

\noindent\textbf{Hyperparameter Ablation.} To validate the robustness of the trade-off parameter in Eq. (7), we rewrite Eq. (7) as: $\mathcal{L} = \lambda \cdot \mathcal{L}_{\text{hyper}} +  \mathcal{L}_{\text{rec}}$, ablate the loss to analyze this trade-off. As shown in \cref{fig: hyperparameter}, increasing $\lambda$ improves semantic consistency (lower FID) while reducing diversity (lower LPIPS), validating the controllability introduced by the hyperbolic component.
 \begin{table}
\centering

    \resizebox{1\linewidth}{!}{
    \setlength{\abovecaptionskip}{-0.2cm}   
    \setlength{\belowcaptionskip}{-0.5cm}   
    \begin{tabular}{c|ccccc}
        \toprule
        Hyp Radius &  6.2 & 6.0 & 5.5 & 5.0 & 4.5\\
        \midrule
        \textbf{FID($\downarrow$)} & 27.89 & 24.67 & \textbf{23.96} & 24.89 & 26.63\\
        \textbf{LPIPS($\uparrow$)} & 0.7585 & 0.7589 & 0.7595 & 0.7643 & \textbf{0.7725}\\
        \bottomrule
    \end{tabular}
    }
    \caption{\textbf{Ablation study} of different radii on Flowers.}
    \label{tab: ablation_radii_supp}
\end{table}

 \begin{table}
\centering

    \resizebox{1\linewidth}{!}{
    \setlength{\abovecaptionskip}{-0.2cm}   
    \setlength{\belowcaptionskip}{-0.5cm}   
    \begin{tabular}{c|ccccc}
        \toprule
        CFG &  1.0 & 1.1 & 1.3 & 1.5 & 1.7\\
        \midrule
        \textbf{FID($\downarrow$)} & 25.52 & 24.89 & \textbf{23.96} & 26.04 & 25.63 \\       \textbf{LPIPS($\uparrow$)} & 0.7391 & 0.7534 & 0.7595 & 0.7660 & \textbf{0.7737}\\
        \bottomrule
    \end{tabular}
    }
    \caption{\textbf{Ablation study} of the influence of CFG on Flowers.}
    \label{tab: ablation_cfg}
\end{table}

\begin{table}
\centering

    \resizebox{1\linewidth}{!}{
    \setlength{\abovecaptionskip}{-0.2cm}   
    \setlength{\belowcaptionskip}{-0.5cm}   
    \begin{tabular}{c|ccccc}
        \toprule
        Strength & 1.0 & 0.98 & 0.95 & 0.9 & 0.8\\
        \midrule
        \textbf{FID($\downarrow$)} & 28.97 & 24.59 & \textbf{23.96} & 24.94 & 26.48\\
        \textbf{LPIPS($\uparrow$)} & \textbf{0.7631} & 0.7606 & 0.7595 & 0.7585 & 0.7375\\
        \bottomrule
    \end{tabular}
    }
    \caption{\textbf{Ablation study} of the influence of encoding strength on Flowers.}
    \label{tab: ablation_strength}
\end{table}

\begin{figure}[t]
  \centering
  \setlength{\abovecaptionskip}{0.0cm}   
  \setlength{\belowcaptionskip}{-0.2cm}   
   \includegraphics[width=1.0\linewidth]{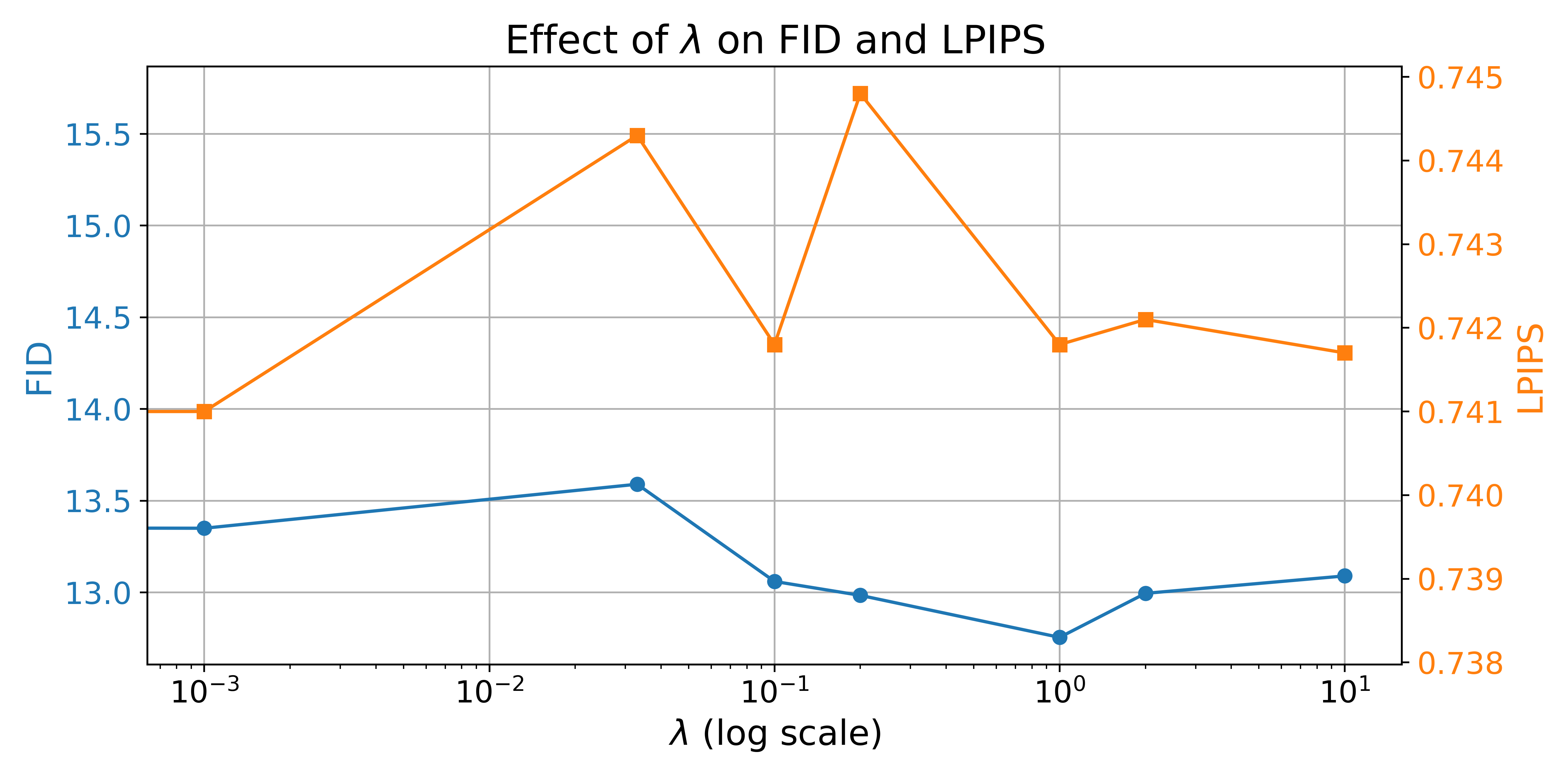}

   \caption{\textbf{Ablation Study} of trade-off adaptive hyperparameter $\lambda$ in the loss function.}
   \label{fig: hyperparameter}
\end{figure}

\begin{figure*}[h]
    \centering
    \includegraphics[width=1.0\textwidth]{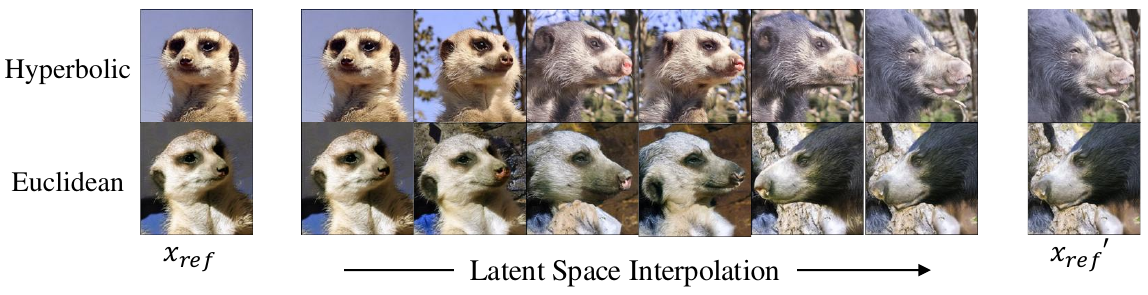}
    \vspace{0ex}
    \caption{\textbf{Comparison of interpolation in hyperbolic space and Euclidean space} on Animal Faces dataset.}
    \label{fig: interpolation-comparison-2}
\end{figure*}

\begin{figure*}[h]
    \centering
    \includegraphics[width=1.0\textwidth]{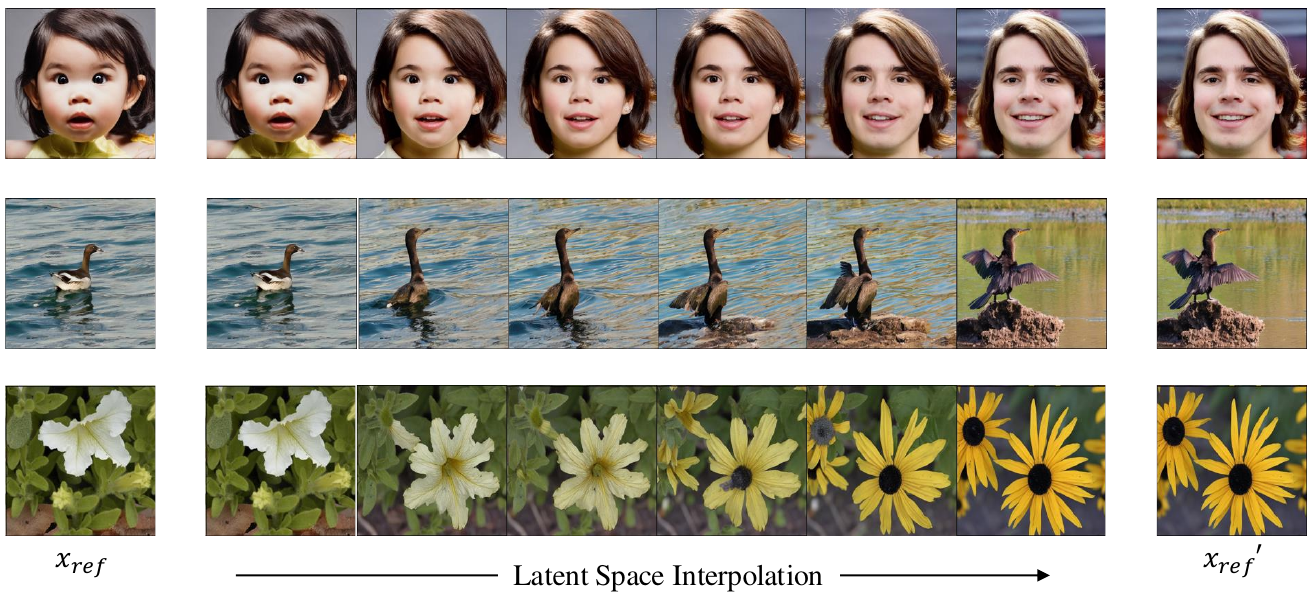}
    \vspace{0ex}
    \caption{\textbf{More results of interpolation in hyperbolic space} on FFHQ, NABirds, and Flowers datasets.}
    \label{fig: interpolation-more}
\end{figure*}

\section{Comparison with Euclidean space}
\label{appendix: Comparison with Euclidean space}
In this section, we present a detailed comparison of different latent spaces, as shown in \cref{fig: interpolation-comparison-2}. Compared to classical Euclidean space, hyperbolic space enables smoother transitions between two given images. In hyperbolic space, identity-irrelevant features transition first, followed by a gradual change in identity-relevant features. In contrast, Euclidean space exhibits simultaneous changes in both identity-relevant and identity-irrelevant features, leading to less structured transitions.

These results confirm that our method effectively learns hierarchical representations in hyperbolic space, enabling few-shot image generation by selectively modifying category-irrelevant features—a capability that Euclidean space cannot achieve. Additional interpolation results, provided in \cref{fig: interpolation-more}, demonstrate that smooth and distortion-free transitions are achievable in hyperbolic space. These findings highlight that \modelname{} enables precise geodesic and hierarchical control during editing, offering a significant advantage over traditional approaches.

\section{Out-of-distribution Few-shot Image Generation}
\label{appendix: Out-of-distribution Few-shot Image Generation}
\begin{figure*}[h]
    \centering
    \includegraphics[width=1.0\textwidth]{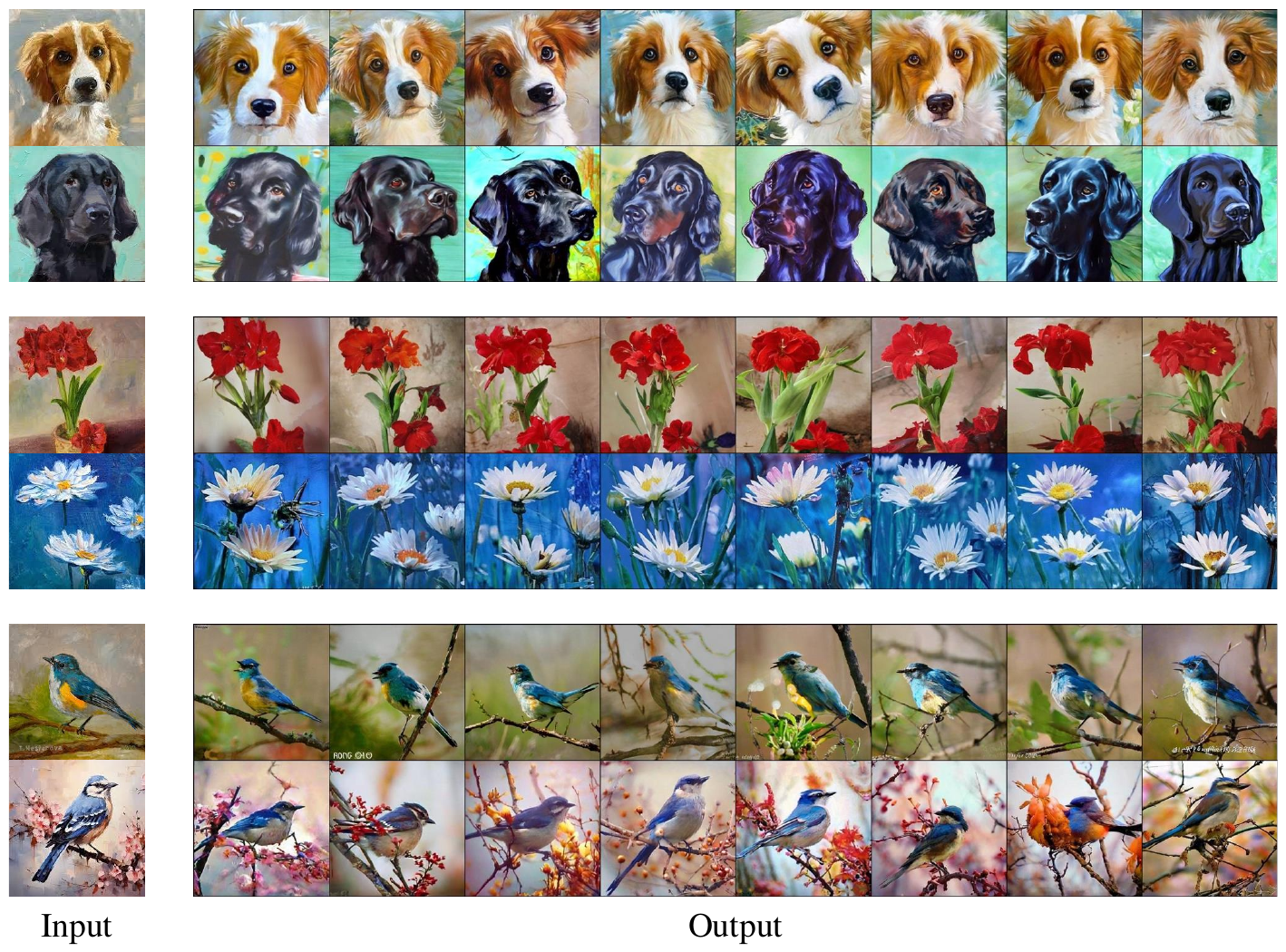}
    \vspace{0ex}
    \caption{Few-shot image generation on \textbf{out-of-distribution examples in painting style}.}
    \label{fig: ood_painting}
\end{figure*}

\begin{figure*}[h]
    \centering
    \includegraphics[width=1.0\textwidth]{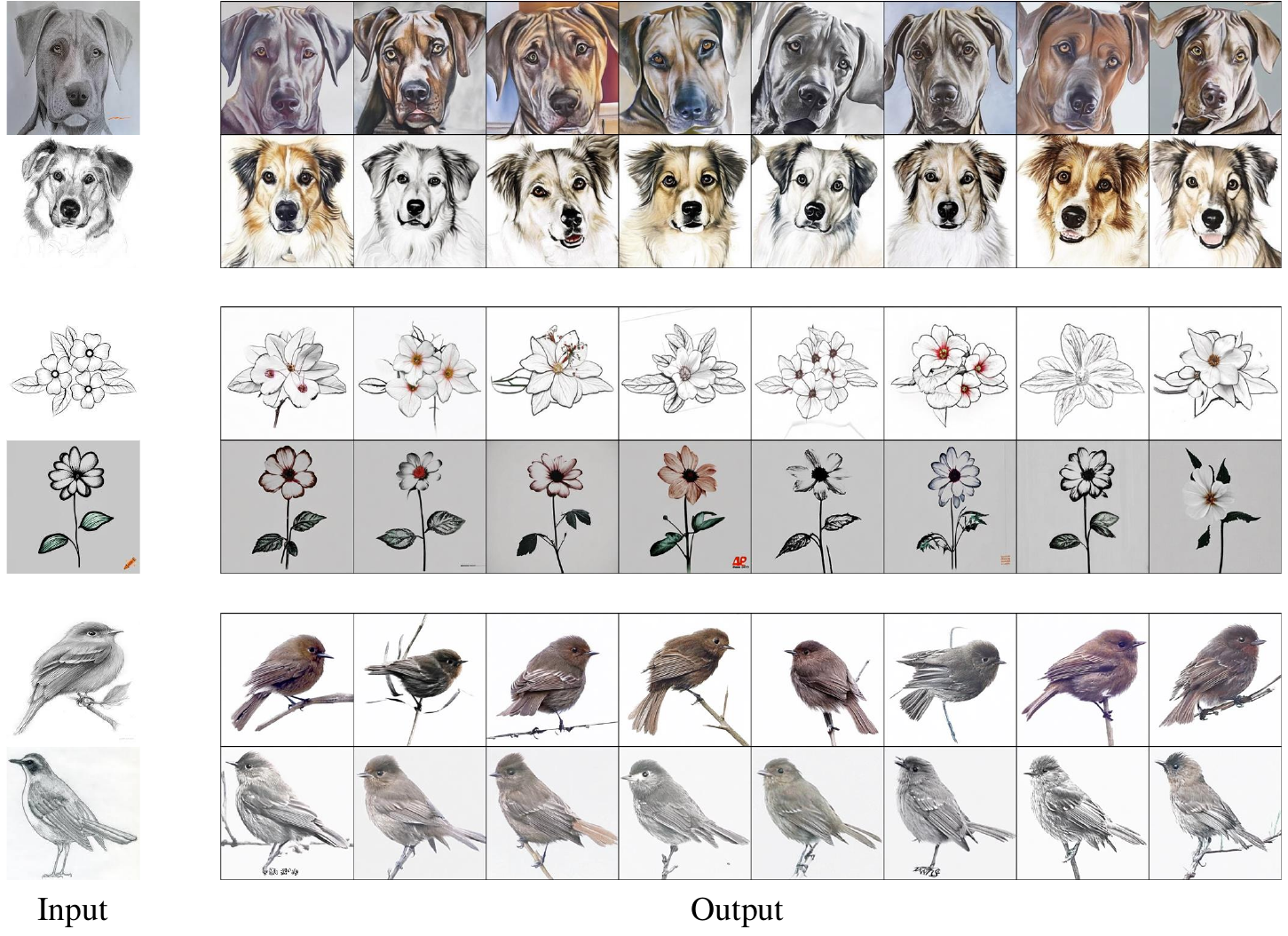}
    \vspace{0ex}
    \caption{Few-shot image generation on \textbf{out-of-distribution examples in sketch style}.}
    \label{fig: ood_sketch}
\end{figure*}

\begin{figure*}[h]
    \centering
    \includegraphics[width=1.0\textwidth]{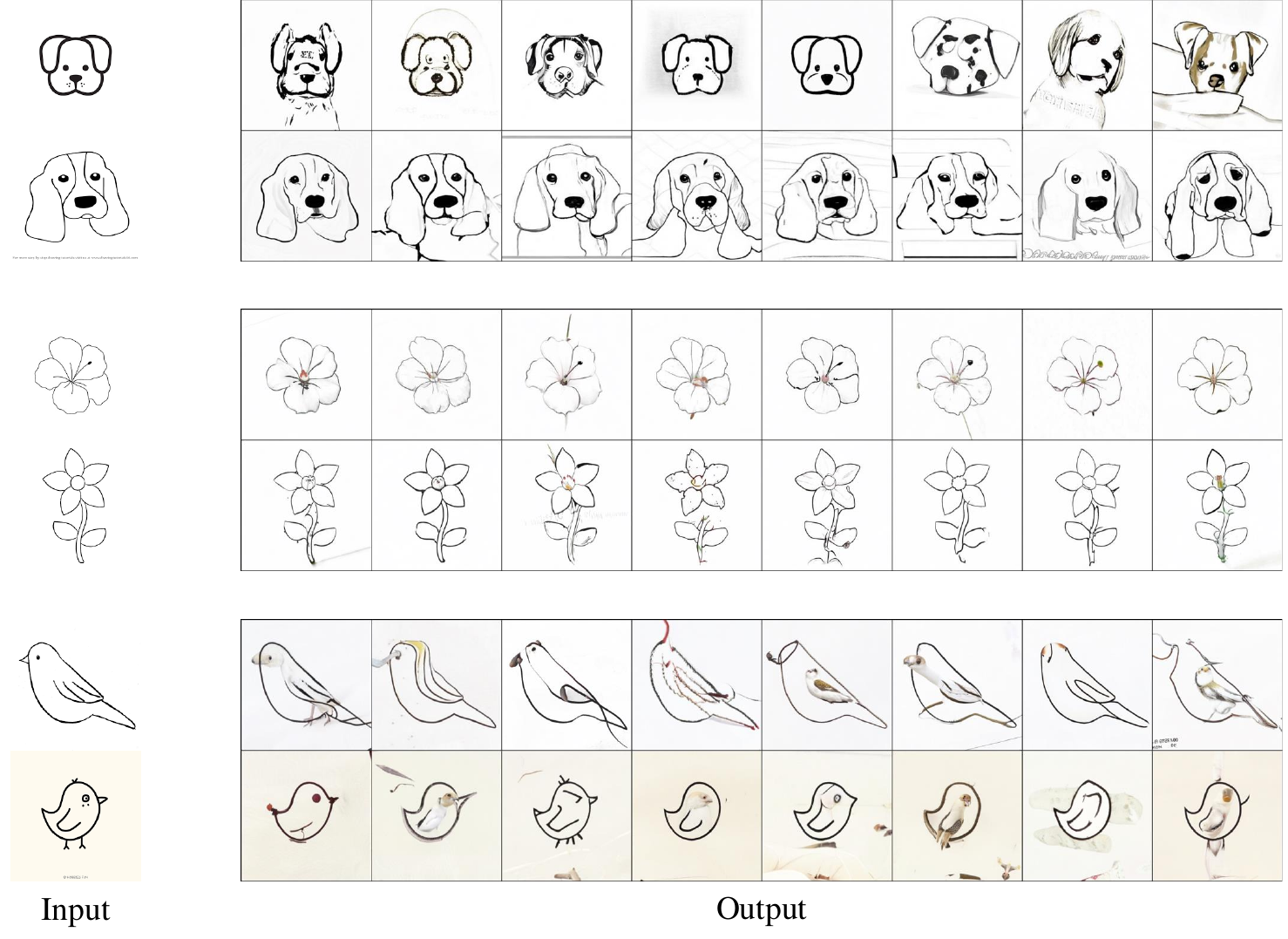}
    \vspace{0ex}
    \caption{Few-shot image generation on \textbf{out-of-distribution examples in quick draw style}.}
    \label{fig: ood_quickdraw}
\end{figure*}

\begin{figure*}[h]
    \centering
    \includegraphics[width=1.0\textwidth]{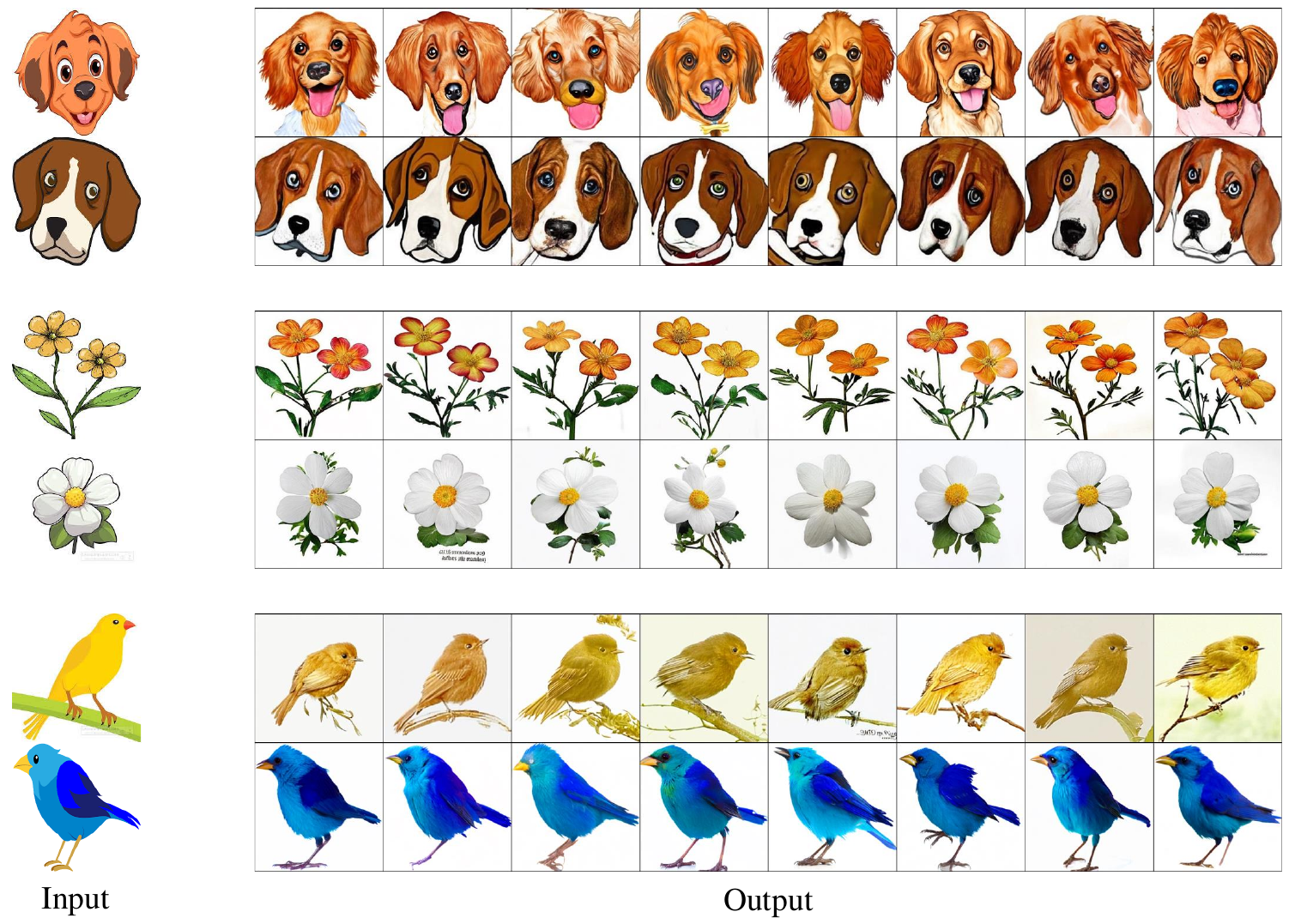}
    \vspace{0ex}
    \caption{Few-shot image generation on \textbf{out-of-distribution examples in clipart style}.}
    \label{fig: ood_clipart}
\end{figure*}

In Sec 4.2 of the main paper, we mentioned we fine-tuned the model trained with VGGFaces using the FFHQ dataset. The model shows exceptional out-of-distribution generalization ability on the FFHQ dataset. To further verify the OOD generalization ability of \modelname{}, we select two images for Animal Faces~\cite{Liu19Few}, Flowers~\cite{Nilsback08}, and NABirds~\cite{Horn15Nabirds} datasets with three styles from DomainNet~\cite{Peng19DomainNet} including `` painting'', ``sketch'', ``quick draw'', and ``clipart'' styles where are model never seen during the training stage. The OOD style transfer can be done by slightly increasing the encoding strength of the stochastic encoder to capture more style information of the given new images. The results in~\cref{fig: ood_painting}~\cref{fig: ood_sketch}~\cref{fig: ood_quickdraw}~\cref{fig: ood_clipart} show that our proposed method has exceptional OOD generalization ability even for new domains with a big gap from the original domain. Although our model still generates images with some real detail for the style ``clipart'', the performance in other styles is satisfying. Such an OOD generalization ability is significantly better than any of the previous works.

\section{Hierarchical Image Generation}
\label{appendix: Hierarchical Image Generation}
\begin{figure*}[h]
    \centering
    \includegraphics[width=0.9\textwidth]{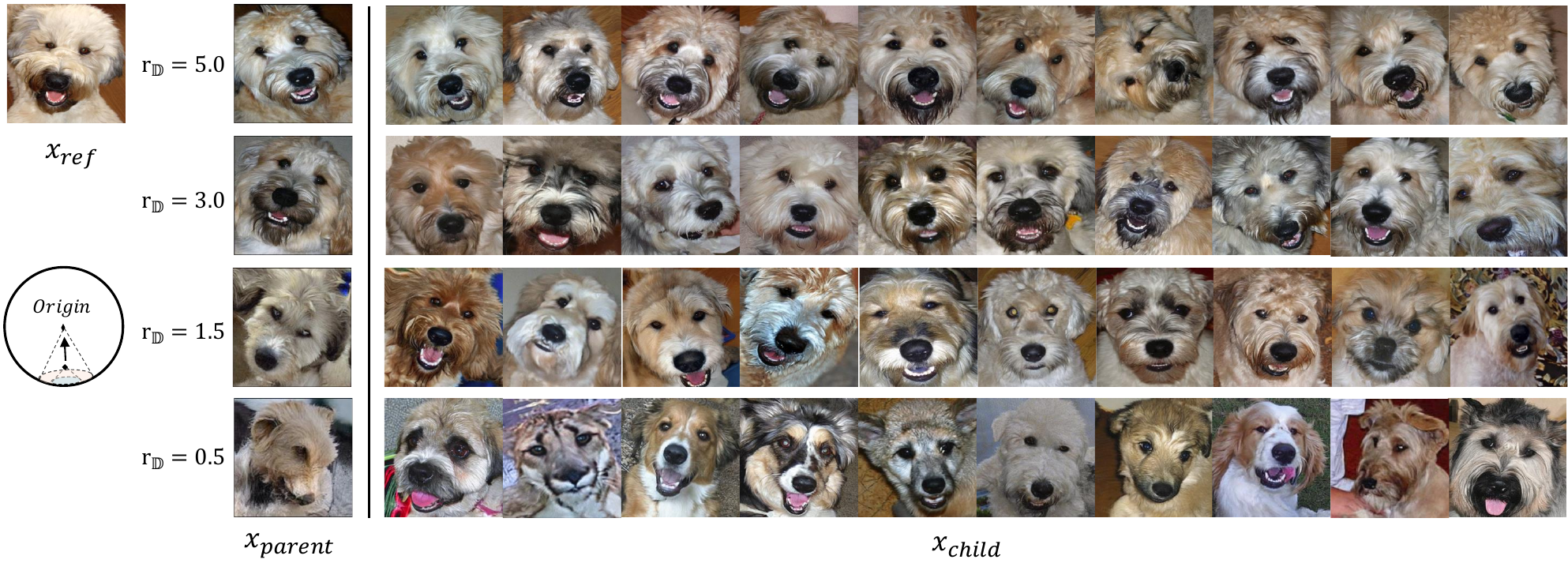}
    \caption{\textbf{Images with hierarchical semantic similarity generated by \modelname{}}.}
    \label{fig: hierarchy_animals}
\end{figure*}

\begin{figure*}[h]
    \centering
    \includegraphics[width=0.9\textwidth]{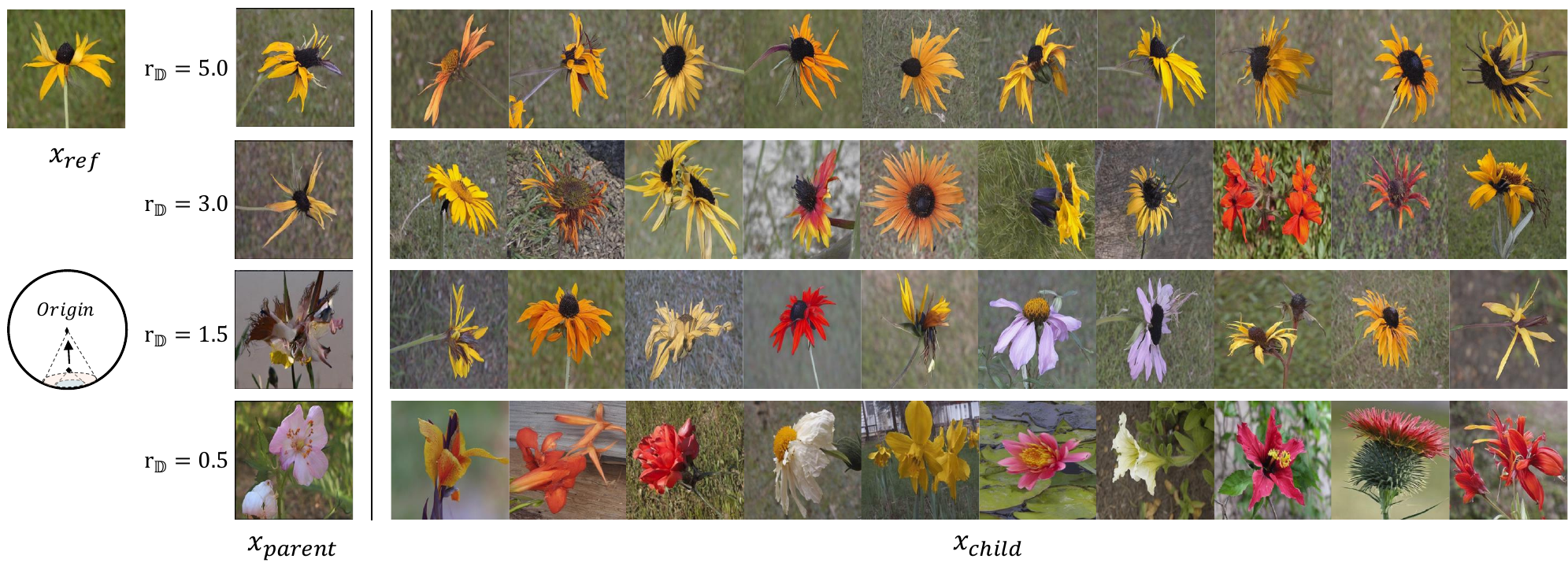}
    \caption{\textbf{Images with hierarchical semantic similarity generated by \modelname{}}.}
    \label{fig: hierarchy_flowers}
\end{figure*}

\begin{figure*}[h]
    \centering
    \includegraphics[width=0.9\textwidth]{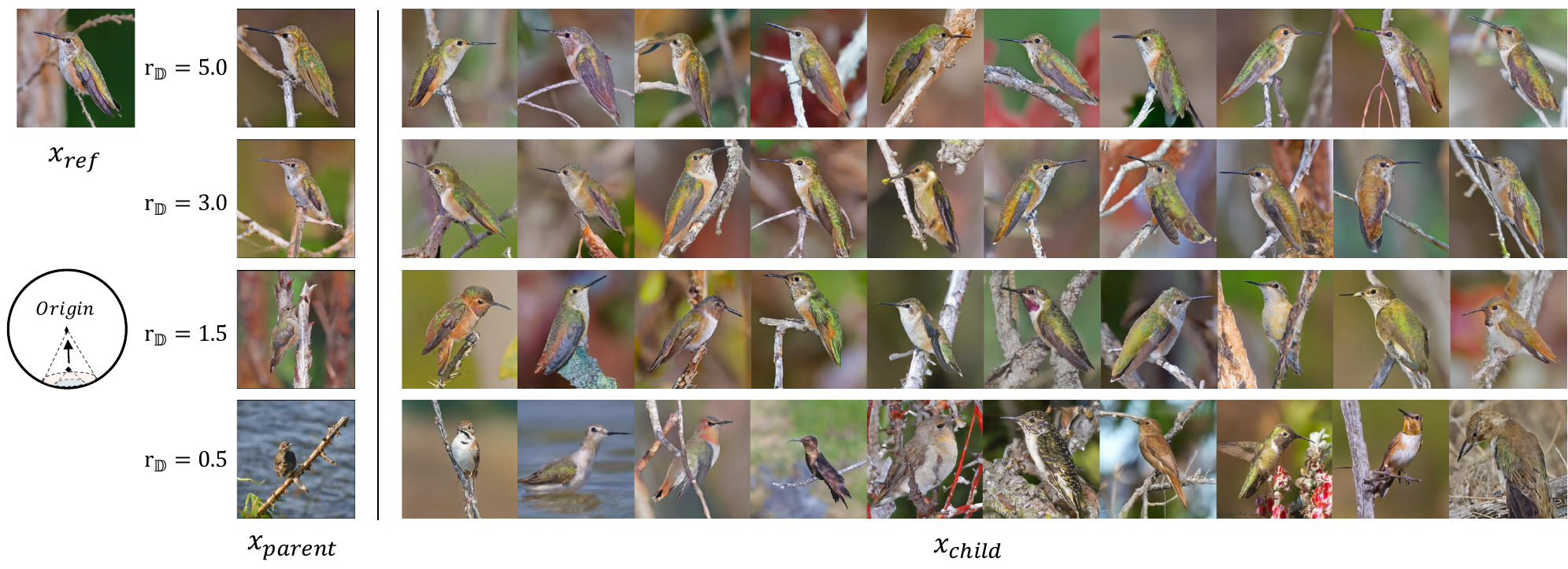}
    \caption{\textbf{Images with hierarchical semantic similarity generated by \modelname{}}.}
    \label{fig: hierarchy_birds}
\end{figure*}

In this section, we provide additional examples of images generated by \modelname{} at varying radii in the Poincaré disk. As illustrated in \cref{fig: hierarchy_animals},~\cref{fig: hierarchy_flowers}, and~\cref{fig: hierarchy_birds}, high-level, category-relevant attributes remain unchanged when the radius is large, allowing for the generation of diverse images within the same category. Conversely, as the hyperbolic radius $r_\mathbb{D}$ decreases, the generated images become more abstract and semantically diverse. Moving closer to the center of the Poincaré disk results in the gradual loss of fine-grained details and changes to higher-level attributes.

For the few-shot image generation task, larger radii are optimal as they allow for the modification of category-irrelevant attributes while preserving the category identity. However, \modelname{} is not limited to few-shot image generation and shows significant potential for other downstream applications. For example, \modelname{} can generate a diverse set of feline images from a single cat image. This is achieved by scaling the latent code to a smaller radius in hyperbolic space and introducing random perturbations to approximate the average latent code for various feline categories. Finally, fine-grained and diverse feline images are generated by moving these average codes outward to larger radii, representing their "children" in the hierarchical space.

\section{Comparison with State-of-the-art Few-shot Image Generation Method}
\label{appendix: Comparison with State-of-the-art few-shot image generation Method}
\begin{figure*}[h]
    \centering
    \includegraphics[width=1.0\textwidth]{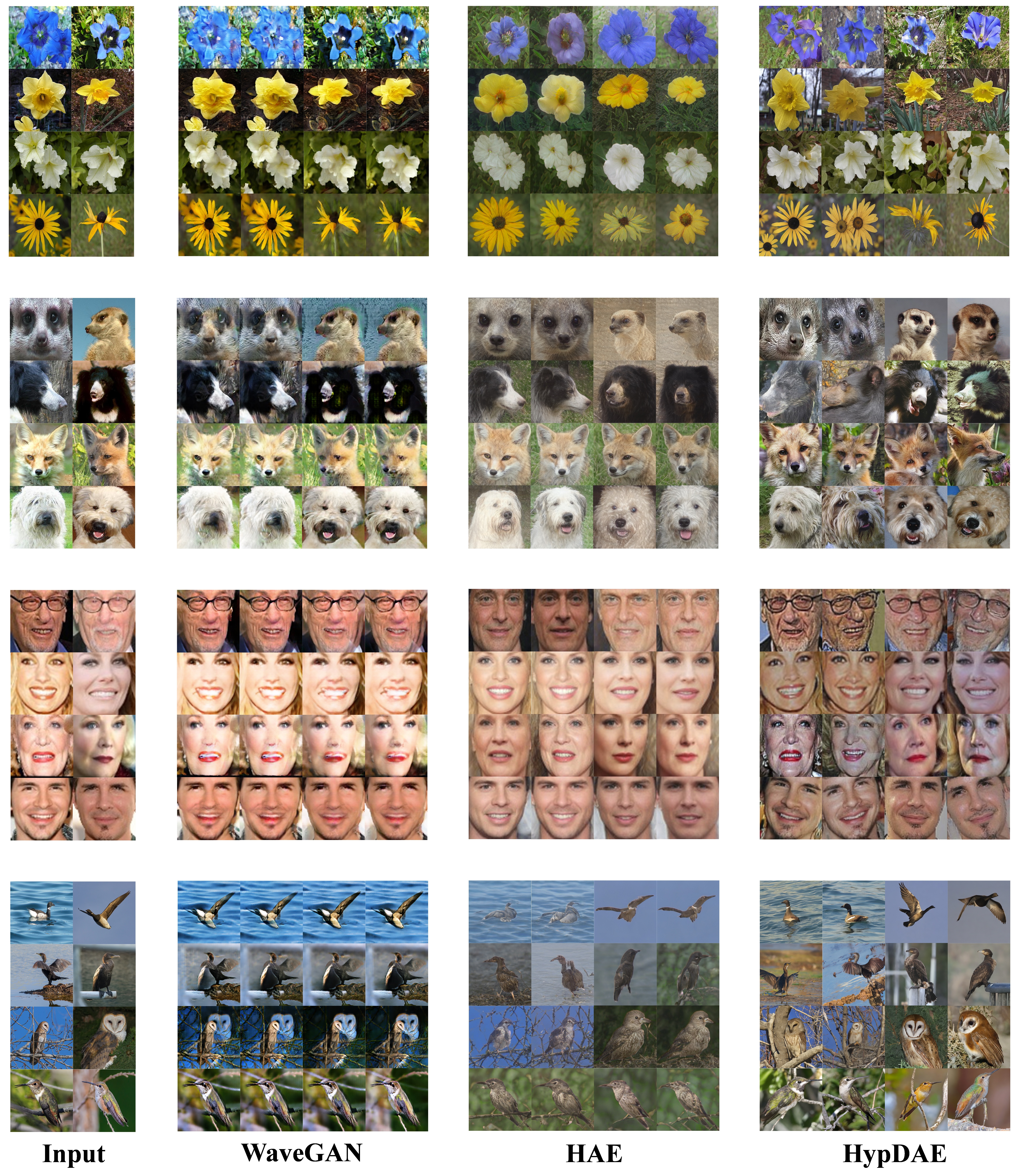}
    \vspace{0ex}
    \caption{\textbf{More comparison between images generated by WaveGAN, HAE, and \modelname{}} on Flowers, Animal Faces, VGGFaces, and NABirds. Note: WaveGAN uses a 2-shot setting; HAE and \modelname{} are both in a 1-shot setting. \textbf{Zoom in to see the details.}}
    \label{fig: comparison_more}
\end{figure*}

We compare images generated by state-of-the-art methods, including WaveGAN~\cite{yang22wavegan}, HAE~\cite{Li23HAE}, and our proposed method, across four datasets. As shown in \cref{fig: comparison_more}, WaveGAN produces high-fidelity images, but the diversity is limited (\ie, blending features from two input images without significant variation). HAE improves diversity but suffers from low fidelity and quality, with missing details and changes in category or identity compared to the original images. In contrast, \modelname{} achieves an excellent balance between maintaining identity and enhancing diversity while delivering significantly higher image quality than other methods. These results highlight the potential of \modelname{} for broader applications in future downstream tasks.

\section{User Study}
\label{appendix: User Study}
\begin{figure*}[h]
    \centering
    \includegraphics[width=1.0\textwidth]{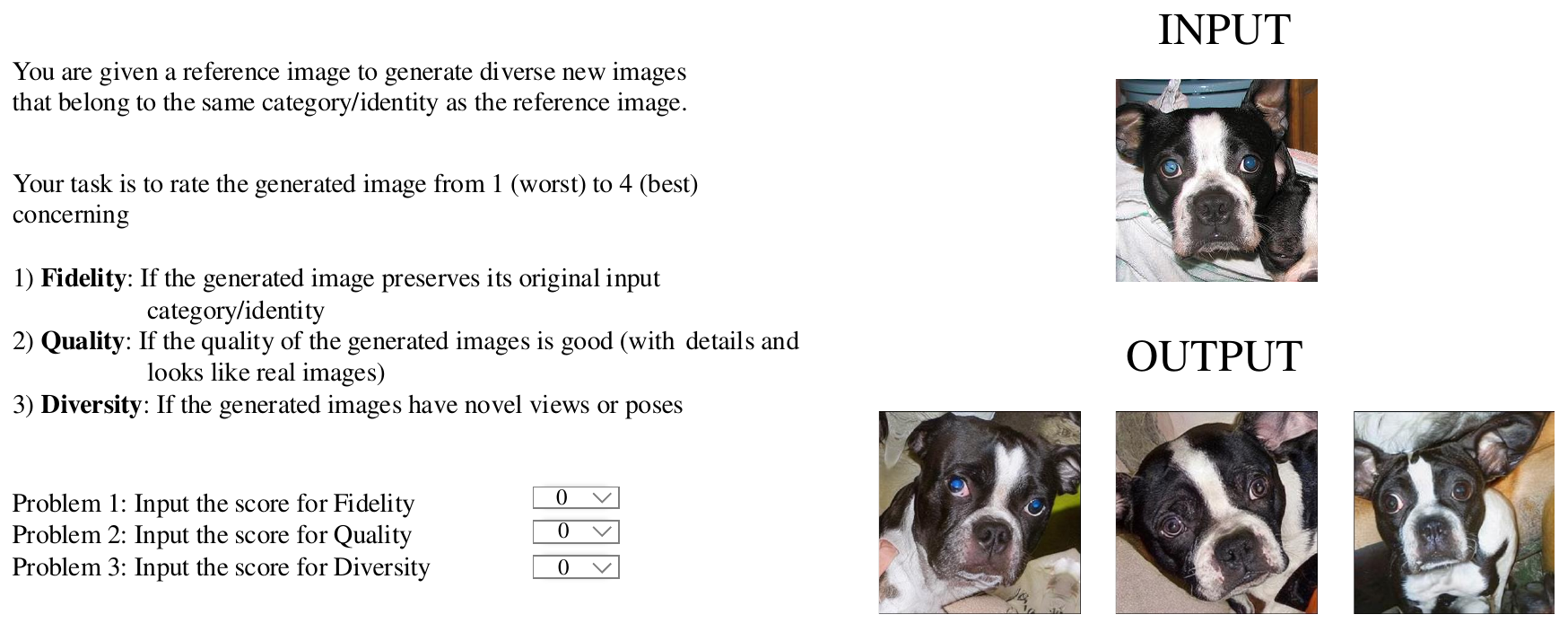}
    \vspace{0ex}
    \caption{\textbf{The illustration of the user study interface.}}
    \label{fig: human-study-interface}
\end{figure*}

As mentioned, we conducted an extensive user study with a fully randomized survey. Results are shown in the main text. Specifically, we compared \modelname{} with three other models WaveGAN~\cite{yang22wavegan}, HAE~\cite{Li23HAE}, and SAGE~\cite{Ding23}:
\begin{enumerate}
    \item We randomly chose 5 images from four datasets, and for each image, we then generated 3 variants in 1-shot setting (WaveGAN used 2-shot setting), respectively. Overall, there were 20 original images and 60 generated variants in total.
    \item For each sample of each model, we present one masked background image, a reference object, and the generated image to annotators. We then shuffled the orders for all images.
    \item We recruited 30 volunteers from diverse backgrounds and provided detailed guidelines and templates for evaluation. Annotators rated the images on a scale of 1 to 4 across three criteria: ``Fidelity'', ``Quality'', and ``Diversity''. ``Fidelity'' evaluates identity preservation, while ``Quality'' assesses quality of the images (\eg, details of the image). ``Diversity'' measures variation among generated proposals to discourage ``copy-paste'' style outputs.
\end{enumerate}
The user-study interface is shown in \cref{fig: human-study-interface}.

\section{Additional Examples Generated by \modelname{}}
\label{appendix: Additional Examples Generated by HypDAE}
\begin{figure*}[h]
    \centering
    \includegraphics[width=1.0\textwidth]{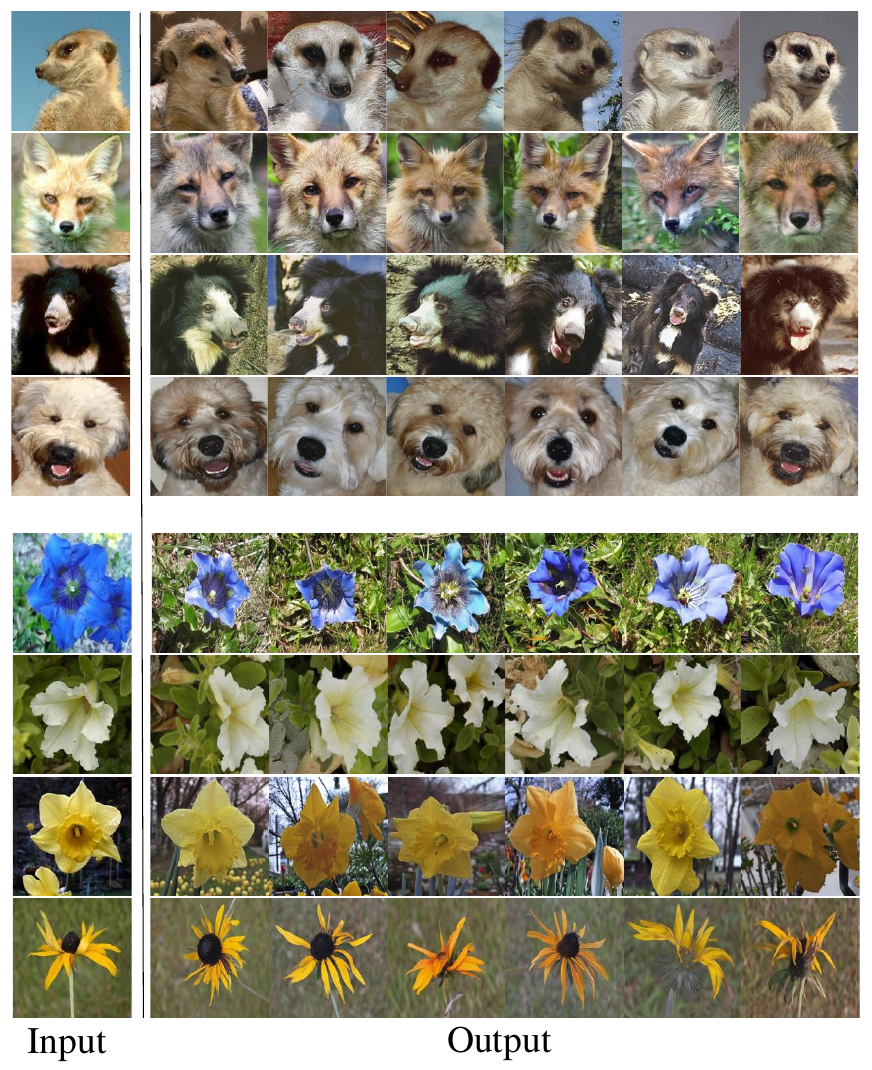}
    \vspace{0ex}
    \caption{\textbf{More examples generated by \modelname{}} on Animal Faces and Flowers.}
    \label{fig: more-generation-1}
\end{figure*}

\begin{figure*}[h]
    \centering
    \includegraphics[width=1.0\textwidth]{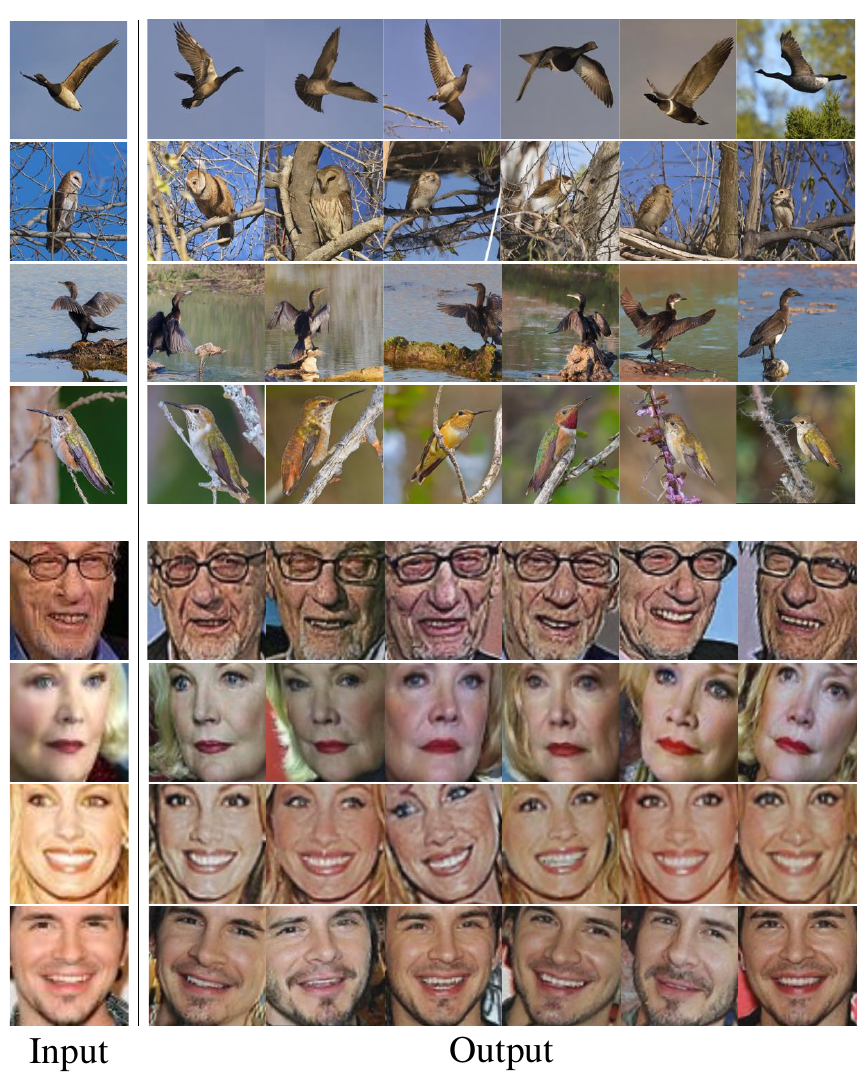}
    \vspace{0ex}
    \caption{\textbf{More examples generated by \modelname{}} on NABirds and VGGFaces.}
    \label{fig: more-generation-2}
\end{figure*}

Finally, we provide more examples generated by \modelname{} in~\cref{fig: more-generation-1} and~\cref{fig: more-generation-2} for four datasets. The results show that our method achieves a balance between the quality and diversity of the generated images which significantly outperforms previous methods.

\clearpage

\end{document}